\documentclass{article}
\usepackage{enumitem}
\usepackage{booktabs}
\usepackage{xcolor,capt-of}
\usepackage{tikz}
\usepackage[export]{adjustbox}
\usepackage{float}
\definecolor{insightone}{RGB}{30, 120, 210}     
\definecolor{insighttwo}{RGB}{240, 190, 20}     
\definecolor{insightthree}{RGB}{40, 170, 40}    
\definecolor{insightfour}{RGB}{240, 190, 20}    

\DeclareRobustCommand{\insightbadge}[3][70]{%
  \tikz[baseline=(badge.base)]\node[
    circle,
    fill=#2!#1,
    text=white,
    font=\bfseries\footnotesize,
    inner sep=0.5pt,
    minimum size=1.5em,
    anchor=base
  ] (badge) {#3};%
}

\newcommand{\Ione}{\protect\insightbadge{insightone}{R1}}
\newcommand{\Itwo}{\protect\insightbadge{insighttwo}{R2}}
\newcommand{\Ithree}{\protect\insightbadge{insightthree}{R3}}

\usepackage{soul}
\sethlcolor{insightone!100}  

\usepackage[preprint]{neurips_2026}

\usepackage[utf8]{inputenc} 
\usepackage[T1]{fontenc}    
\usepackage{url}            
\usepackage{booktabs}       
\usepackage{amsfonts}       
\usepackage{nicefrac}       
\usepackage{microtype}      
\usepackage{xcolor}         
\usepackage{amsmath} 
\usepackage{graphicx} 
\usepackage{multirow}
\usepackage{hyperref}
\usepackage{makecell}
\usepackage{enumitem}

\usepackage{tcolorbox}
\usepackage{listings}

\tcbuselibrary{listings, skins, breakable}

\newtcblisting{promptbox}[1]{
    colback=gray!8,
    colframe=gray!40,
    fonttitle=\small\bfseries,
    title=#1,
    listing only,
    listing options={
        basicstyle=\small\ttfamily,
        breaklines=true,
        breakatwhitespace=false,
        columns=fullflexible,
        keepspaces=true,
    },
    left=4pt, right=4pt, top=4pt, bottom=4pt,
    breakable,
}

\title{Building Fast, Evaluating Slow: Pipeline Choices Dominate Autointerpretability Score Variance}

\usepackage{microtype}
\begin{document}

\author{
  Sinie van der Ben \\
  ETH Zürich \\
  \And
  Neele Roch \\
  ETH Zürich \\
  \AND
  Anna Hedström \\
  ETH Zürich \\
  \AND
  Mennatallah El-Assady \\
  ETH Zürich \\
}

\maketitle


\begin{abstract} 
    Cross-paper comparison of sparse autoencoder (SAE) interpretability often relies on autointerpretability scores. In this evaluation pipeline, a language model (LM) explains each feature, and another LM scores the explanation. For these comparisons to be meaningful, scores must reflect stable properties of the features rather than confounding aspects of the evaluation pipeline. Through systematic experiments across four metrics (simulation, detection, fuzzing, purity), two models (Pythia-160M, Apertus-8B), and four axes of methodological variation, we show that this assumption does not hold. Specifically, we find that \textbf{\textcolor{insightone}{(R1)}} methodological variance collectively exceeds architectural variance across all metrics and tested models; \textbf{\textcolor{insighttwo}{(R2)}} each metric exhibits a distinct instability profile, with detection being the most stable and fuzzing unreliable across all conditions; \textbf{\textcolor{insightthree}{(R3)}} top-$k$ feature rankings do not stay consistent across corpus and draw conditions, masking per-feature instability behind stable mean scores; a failure that cannot be detected by monitoring explanation similarity alone. These findings suggest that cross-paper comparisons based on autointerpretability scores may reflect pipeline differences rather than architectural differences, with implications for the ongoing debate on SAE utility. More broadly, unreliable evaluation slows progress in interpretability research at a time when reliable tools for understanding AI systems are needed. To support evaluation, we contribute a variance decomposition approach, a Stability Check, and a Minimum Reporting Checklist.
\end{abstract}

\begin{figure}[h]
\includegraphics[width=1\textwidth]{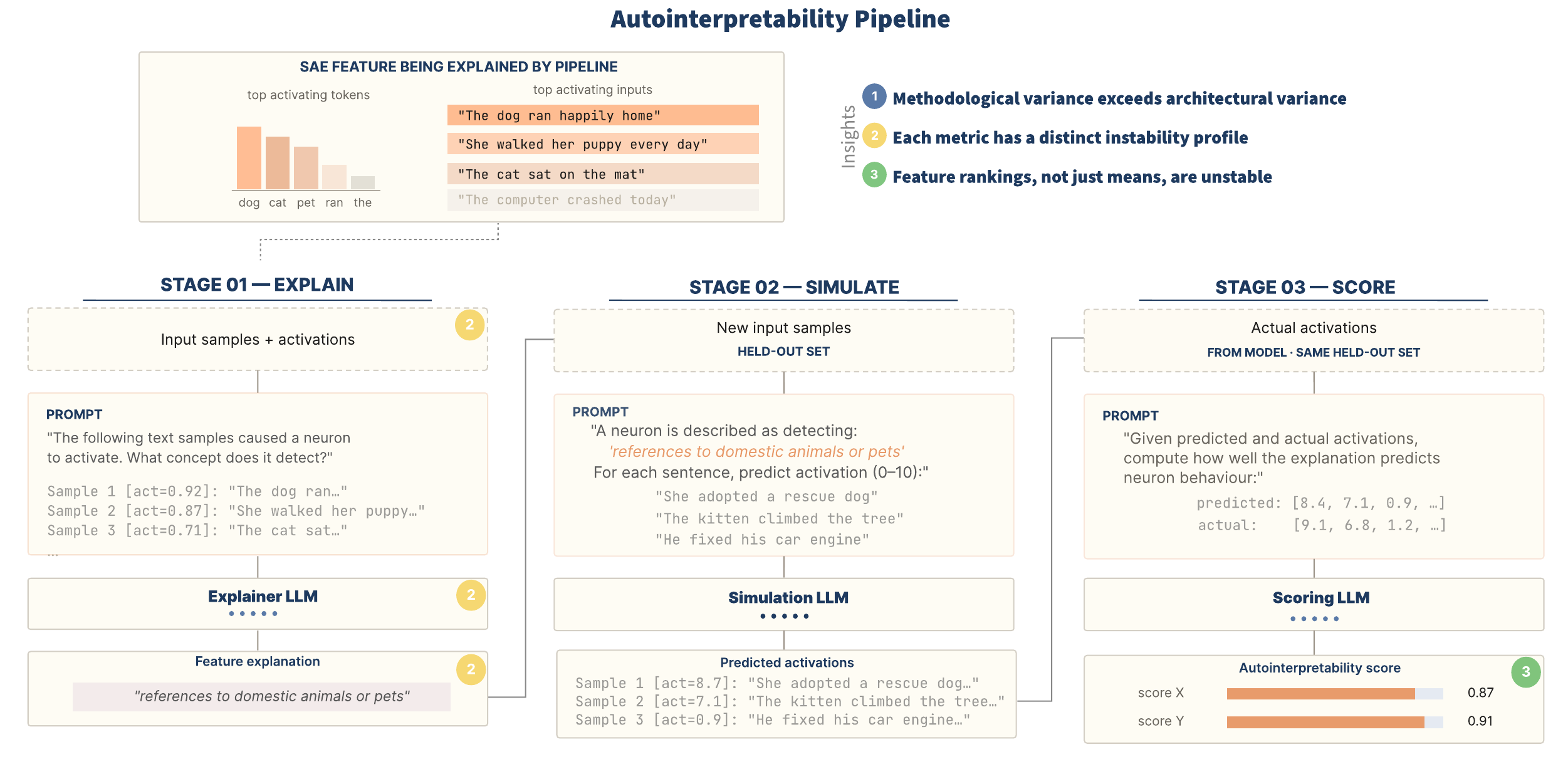}
\centering
\caption{Overview of Automated Interpretability Pipeline with the insights location. \textbf{\textcolor{insightone}{(R1)}} is not shown in the figure, but this insight applies to the pipeline as a whole: methodological variance exceeds architectural variance. {\textcolor{insighttwo}{(R2)}} shows that each metric exhibits a distinct instability profile, while it is shown through different perturbations of corpora, draws, explainers, and paraphrases. {\textcolor{insightthree}{(R3)}} shows variability in the rankings of feature scores.}
\vspace{-1em}
 \label{fig:overview_auto}
\end{figure}

\section{Introduction}

Autointerpretability scores are widely relied upon as the main instrument for feature interpretability assessment in SAEs \cite{bricken2023monosemanticity, cunningham2024sparse, juang2024opensource, templeton2024scaling}, and are additionally used in architecture comparisons \cite{bussmann2025learning, karvonen2025saebench}. However, if these scores reflect evaluation pipeline configurations rather than feature interpretability, such comparisons are potentially meaningless. Proxy metrics such as reconstruction error and sparsity reflect optimization behavior rather than semantic quality \cite{chanin2025sparse, minegishi2025rethinking}, and SAEs that perform well on these metrics have been shown to detect only a small subset of true features \cite{korznikov2026sanity}. This has motivated evaluation frameworks that target interpretability more directly \cite{makelov2025towards, paulo2025automatically, puri2025fade}, yet a fundamental question remains unanswered: \textit{``\textbf{Are current autointerpretability evaluation practices stable enough to support cross-paper comparisons, or do they reflect pipeline choices?}''}

In autointerpretability, or automated interpretability, a language model (LM) 
generates a natural language description of a feature, and a second LM 
evaluates how well that description predicts feature activations on held-out 
examples \cite{bills2023language}. This pipeline currently underlies most SAE interpretability evaluations \cite{cunningham2024sparse, karvonen2025saebench, templeton2024scaling}. Prior work has identified individual sources of pipeline variation that affect scores, such as sample selection strategies \cite{juang2024opensource, paulo2025automatically}, explainer or scorer model capacity \cite{juang2024opensource, paulo2025automatically, puri2025fade}, and prompt construction \cite{puri2025fade}, but no study has systematically quantified their \textit{relative} contribution to total score variance, or assessed whether their combined effect exceeds the variance attributable to 
architectural differences between SAEs.

After the introduction of SAEs 
\cite{bricken2023monosemanticity, cunningham2024sparse}, they have been adopted fast in the field of interpretability. Early work demonstrates that identified features can be used to successfully steer model behavior \cite{templeton2024scaling}, while other applications include detecting and mitigating emergent misalignment \cite{wang2025persona}. Simultaneously, 
SAE features are sensitive to small input perturbations \cite{li2026evaluating}; SAEs trained on identical data learn different features across random seeds \cite{paulo2025sparseautoencoderstraineddata}, and downstream task performance does not consistently exceed simpler baselines \cite{kantamneni2025are, wu2025axbench}. Whether SAEs deliver on interpretability remains an important question for the debate on their usefulness, and that debate requires reliable metrics to resolve.

In this paper, we show that current dominant instruments are unreliable. 
Through systematic experiments on two models (\texttt{Pythia-160M}, 
\texttt{Apertus-8B}) across four metrics (simulation, detection, fuzzing, 
purity; \autoref{sec:metrics}), we vary four axes of the evaluation pipeline 
independently and measure instability using ICC \cite{koo2016guideline} and 
Top-$k$ Jaccard \cite{kuncheva2007stability} (\autoref{sec:quant_instability}). 
We study metric \textit{reliability} rather than \textit{validity}, under specific pipeline configurations. We do not evaluate whether autointerpretability scores truly capture interpretability, but whether they are consistent enough to 
support the comparisons the field relies on, which is a precondition for a validity claim.
We make the following contributions:

\begin{enumerate}[label={}]
    \item[\Ione] \textbf{Methodological variance exceeds architectural variance.} 
    Pipeline choices (corpus, sampling, explainer, and phrasing) 
    collectively account for more score variance than SAE architecture, making     cross-architecture comparisons potentially unreliable without standardization 
    (\autoref{sec:findings_overview}).

    \item[\Itwo] \textbf{Each metric has a distinct instability profile.}
    \textsc{Simulation} and \textsc{Purity} are most sensitive to random draw; 
    \textsc{Detection} to corpus choice; \textsc{Fuzzing} is unreliable across 
    all conditions. No single metric captures the full instability picture, 
    motivating multi-metric evaluation (\autoref{sec:findings_metrics}), confirming statements of earlier research \cite{puri2025fade}.

    \item[\Ithree] \textbf{Feature rankings show instability beyond mean scores.}
    Mean scores mask per-feature instability, because top-$k$ feature sets identified on one corpus or draw have near-chance overlap with those from another, with direct consequences for any application that selects specific features (\autoref{sec:findings_rankings}).
\end{enumerate}

Additionally, we find that explanation similarity cannot diagnose score 
instability, as similarity scores remain high across all conditions even 
when scores vary substantially.

Based on these findings, we propose a Stability Check and a Minimum Reporting 
Checklist (\autoref{sec:recommendations}) to support researchers in diagnosing 
evaluation reliability before interpreting score differences as architectural 
effects. Our findings do not resolve whether SAEs are useful, but establish 
what is required to answer that question reliably. A survey of SAE evaluation 
practices across the literature is provided in \autoref{app:related_work}.

\section{Preliminaries}
\label{sec:preliminaries}

This section establishes the experimental setup for testing the reliability of autointerpretability evaluation methods. Our goal is to quantify how much the scores depend on pipeline choices, relative to the architectural differences they are meant to capture. To this end, we evaluate the stability of autointerpretability scores across four axes of methodological variation: evaluation \textsc{corpus}, random \textsc{draw} of activating examples, \textsc{explainer} model, and explanation \textsc{paraphrasing}, and one architectural axis, using three SAEs of varying sizes for both models. Full experimental details are in \autoref{app:experiments}.

\paragraph{Models and SAEs.}
Experiments were conducted on SAEs trained on \texttt{Pythia-160M} \cite{biderman2023pythia} and \texttt{Apertus-8B Instruct} \cite{swissai2025apertus}. For \texttt{Pythia-160M}, we used publicly available \texttt{BatchTopK} checkpoints at layer 8 from SAEBench \cite{karvonen2025saebench} with $k = 160$. For the architecture comparison, we additionally evaluate \texttt{Matryoshka} \cite{bussmann2025learning} and \texttt{Standard} SAEs \cite{cunningham2024sparse}. For \texttt{Apertus-8B}, we trained a \texttt{BatchTopK} SAE with $k = 160$ on layer 16; this is the first work to train and evaluate SAEs on \texttt{Apertus}. Training details are provided in \autoref{app:sae_training_details}. We opted for these two models and their SAEs as they vary in size and this allows to exclude the effect of source model size. 

The model used to generate and score explanations is \texttt{Gemini Flash} \cite{gemini2024gemini15}, varied across explainer conditions with \texttt{GPT4o-mini} \cite{openai2024gpt4ocard}. All scoring uses \texttt{Gemini Flash} as the fixed scorer, isolating explainer identity from scorer variation. The effect of the scorer has been studied before \cite{paulo2025automatically, puri2025fade}.

\paragraph{Corpora and Features.}
We use \texttt{Pile} \cite{gao2020pile800} as the primary evaluation corpus to match the SAE training distribution. For the corpus experiment, three additional datasets are included: \texttt{C4} \cite{raffel2023c4}, \texttt{Wikipedia} \cite{wikidump}, and \texttt{GitHub} code \cite{githubcode2022}, spanning a gradient from similar to dissimilar relative to training data. All datasets are sampled to 10,000 sequences of 32 tokens and accessed via \texttt{HuggingFace}; corpus statistics are reported in \autoref{tab:corpora}.
Per experiment, we target approximately 250 active features, sampling 300 candidates for \texttt{Pythia-160M} and 2,000 for \texttt{Apertus-8B} to account for lower activation rates. Feature sampling details are provided in \autoref{app:feature_sampling}.

\paragraph{Autointerpretability Metrics.}
\label{sec:metrics}
We evaluate using four complementary metrics. \textsc{Simulation} and \textsc{fuzzing} operate at token level; \textsc{detection} and \textsc{purity} at the sequence level. All metrics are LLM-scored, with \textsc{detection}, \textsc{fuzzing} and \textsc{purity} having binary scores on different levels. Full metric definitions and scoring prompts are in \ref{app:metric_details}.
\begin{itemize}
    \item \textbf{Simulation} \cite{bills2023language}. An LLM predicts token-level feature activations from the explanation alone. The score is the Pearson correlation 
    between predicted and actual activations, $r \in [-1, 1]$; higher is better, $r = 0$ indicates chance.

    \item \textbf{Detection} \cite{paulo2025automatically}. An LLM classifies sequences as activating or non-activating. The score is AUC over top-activating vs.\ random negatives, ranging from 0 to 1; 0.5 indicates chance.

    \item \textbf{Fuzzing} \cite{paulo2025automatically}. Token-level variant of detection: an LLM classifies whether marked tokens within a sequence correctly represent the feature. Score is AUC over token-level predictions, ranging from 0 to 1.

    \item \textbf{Purity} \cite{puri2025fade}. Measures whether high activations are exclusive to the target concept. A scorer LLM assigns (0,1,2) labels to natural samples, and Average Precision is computed over feature activation magnitudes.
\end{itemize}

To assess the semantic consistency of explanations across conditions, we compute pairwise cosine similarity using \texttt{E5-small-v2} embeddings and \texttt{BERTScore} O1 \cite{zhang2020bertscore}. Details are in \autoref{app:explanation_similarity_method}.

\subsection{Quantifying Instability}
\label{sec:quant_instability}

We quantify within-feature instability using three complementary measures. All three track how much the \emph{same} feature's score changes across conditions rather than differences between features. An overview of the metrics is shown in \autoref{tab:metrics_summary} and more details are in \autoref{app:instability_quant}. 

\paragraph{Score Reliability.}
The \textbf{Intraclass Correlation Coefficient} (ICC) \cite{koo2016guideline, liljequist2019intraclass} 
measures the proportion of total variance attributable to stable between-feature differences (Eq.~\ref{eq:icc}). We generally interpret ICC $\geq 0.75$ as good, 
$0.50$--$0.75$ as moderate, and $< 0.50$ as poor reliability, reported with 95\% confidence intervals. ICC model specification is in \autoref{app:icc_spec}.
\begin{equation}
    \text{ICC} = \frac{\sigma^2_{\text{between}}}{\sigma^2_{\text{between}} + 
    \sigma^2_{\text{within}}} \label{eq:icc}
\end{equation}

\paragraph{Ranking Stability.}
Whether the same top-$k$ features are identified across conditions is measured 
by \textbf{Top-$k$ Jaccard} (Eq.~\ref{eq:jaccard}) \cite{kuncheva2007stability}. We report \emph{lift} over chance so results are comparable across pool sizes; 
lift $= 1$ indicates chance-level selection. Details are in \autoref{app:jaccard_lift}.
\begin{equation}
    J_k(A, B) = \frac{|T_k^A \cap T_k^B|}{|T_k^A \cup T_k^B|} \label{eq:jaccard}
\end{equation}

\paragraph{Variance Decomposition.}
\label{par:variance_decomp}
We fit a random effects model and estimate variance components via restricted maximum likelihood (REML), testing whether methodological variance exceeds architectural variance (Eq.~\ref{eq:hypothesis}). REML caveats are in \autoref{app:variance_decomp}.
We define \emph{methodological variance} as $\sigma^2_{\text{method}} 
= \sigma^2_{\text{corpus}} + \sigma^2_{\text{draw}} + \sigma^2_{\text{explainer}} 
+ \sigma^2_{\text{phrasing}}$ and test the central hypothesis:
\begin{equation}
    \sigma^2_{\text{method}} > \sigma^2_{\text{arch}}
    \label{eq:hypothesis}
\end{equation}

\begin{table}[h!]
    \centering
    \caption{Summary of metrics used to evaluate within-feature instability for the four different scores.}
    \label{tab:metrics_summary}
    \footnotesize 
    \begin{tabular}{llcp{5.5cm}}
        \toprule
        \textbf{Metric} & \textbf{Level} & \textbf{Higher/Lower} & \textbf{Question Answered} \\
        \midrule
        ICC & Per-metric & Higher & Is this metric reliable across conditions? \\
        Top-$k$ Jaccard & Per-condition-pair & Higher &  Will my top-$k$ features change? \\
        Variance Decomp \% & Global & Observational & What contributes more to the variance? \\
        \bottomrule
    \end{tabular}
\end{table}

\section{Dissecting Autointerpretability Instability through Four Findings}
\label{sec:insights}
\begin{table}[h!]
\centering
\caption{ICC with 95\% confidence intervals for each source of variation and metric, for \texttt{Pythia-160M} and 
\texttt{Apertus-8B}. ICC(1,1) used for \textsc{Draw}; ICC(2,1) for all other sources. Interpretation follows \cite{koo2016guideline}: $\geq$0.90 excellent, 
0.75--0.90 good, 0.50--0.75 moderate, $<$0.50 poor.}
\label{tab:icc}
\footnotesize 
\resizebox{\textwidth}{!}{%
\begin{tabular}{llcccc}
\toprule
\textbf{Source} & \textbf{Model} & \textbf{\textsc{Simulation}} 
& \textbf{\textsc{Detection}} & \textbf{\textsc{Fuzzing}} 
& \textbf{\textsc{Purity}} \\
\midrule
\multirow{2}{*}{\textsc{Corpus}} 
  & \texttt{Pythia}  & 0.028 $[-0.018, 0.085]$ & 0.151 $[0.087, 0.207]$ 
  & 0.076 $[0.007, 0.153]$ & 0.122 $[0.063, 0.185]$ \\
  & \texttt{Apertus} & 0.155 $[0.082, 0.230]$ & 0.414 $[0.267, 0.531]$ 
  & 0.018 $[-0.089, 0.124]$ & 0.110 $[0.027, 0.187]$ \\
\midrule
\multirow{2}{*}{\textsc{Draw}}   
  & \texttt{Pythia}  & $-0.021$ $[-0.073, 0.036]$ & 0.178 $[0.126, 0.232]$ 
  & 0.349 $[0.280, 0.414]$ & $-0.010$ $[-0.052, 0.038]$ \\
  & \texttt{Apertus} & $-0.049$ $[-0.094, -0.006]$ & 0.071 $[0.014, 0.132]$ 
  & 0.162 $[0.083, 0.234]$ & $-0.078$ $[-0.109, -0.045]$ \\
\midrule
\multirow{2}{*}{\textsc{Explainer}} 
  & \texttt{Pythia}  & 0.568 $[0.362, 0.728]$ & 0.747 $[0.591, 0.854]$ 
  & 0.295 $[-0.133, 0.647]$ & 0.480 $[0.318, 0.627]$ \\
  & \texttt{Apertus} & 0.435 $[0.250, 0.615]$ & 0.846 $[0.796, 0.886]$ 
  & 0.129 $[-0.125, 0.370]$ & 0.289 $[0.140, 0.436]$ \\
\midrule
\multirow{2}{*}{\textsc{Paraphrase}} 
  & \texttt{Pythia}  & 0.860 $[0.783, 0.930]$ & 0.824 $[0.699, 0.895]$ 
  & 0.240 $[0.034, 0.498]$ & 0.650 $[0.523, 0.753]$ \\
  & \texttt{Apertus} & 0.930 $[0.893, 0.958]$ & 0.926 $[0.888, 0.952]$ 
  & 0.267 $[0.114, 0.426]$ & 0.849 $[0.792, 0.897]$ \\
\bottomrule
\end{tabular}
}
\end{table}

\begin{figure}[t]
\includegraphics[width=0.65\textwidth, center]{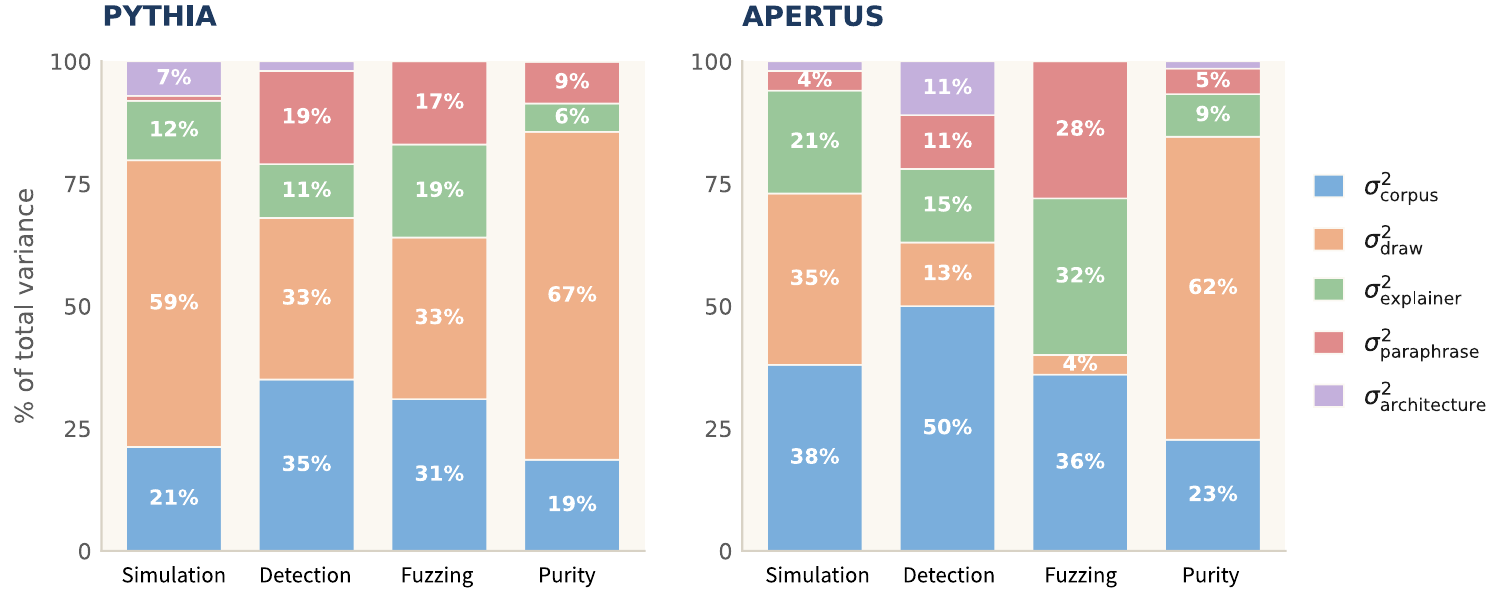}
\centering
\caption{Variance decomposition across five experiments for \texttt{Pythia-160M} and \texttt{Apertus-8B}, with \textsc{Corpus} and \textsc{draw} showing high dominance across most metrics dominates. $\sigma^2_\text{arch}$ remains negligible across all metrics. The full table of percentages can be found in \autoref{app:variance_decomp_table}.}
\vspace{-1em}
\label{fig:variance_decomp}
\end{figure}

\autoref{tab:icc} and \autoref{fig:variance_decomp} summarise the central findings of this paper. Across all four metrics and both models, the variance introduced by pipeline choices,  \textsc{Corpus}, \textsc{Draw}, \textsc{Explainer}, and \textsc{Paraphrase}, collectively exceeds the variance attributable to SAE architecture. At the same time, no two metrics fail in the same way: they fail for different conditions, and the failure modes are invisible relying on mean scores alone \autoref{tab:mean_scores}. Critically, explanation similarity metrics such as cosine similarity and BERTScore \cite{zhang2020bertscore} cannot serve as a diagnostic for this instability, as they remain high even when scores vary substantially (\autoref{app:similarity_values} and \autoref{tab:similarity}). We organize our findings around three insights in the next sections.

\subsection*{\texorpdfstring{\Ione}{R1}~{Methodological Variance Exceeds Architectural Variance}}
\addcontentsline{toc}{subsection}{Methodological Variance Exceeds Architectural Variance}
\label{sec:findings_overview}

\textit{Pipeline choices account for more score variance than architecture across all metrics and both models.}

\autoref{fig:variance_decomp} shows that $\sigma^2_\text{arch}$ contributes negligibly to total score variance on \texttt{Pythia-160M} across all metrics, and remains small on \texttt{Apertus-8B} with a peak of 11\% for \textsc{Detection}, where \textsc{Corpus} alone contributes 50\%. Our central hypothesis $\sigma^2_\text{method} > \sigma^2_\text{arch}$ holds for every metric on both models. This is not a marginal result: for \textsc{Simulation} on \texttt{Pythia-160M}, \textsc{Draw} alone accounts for 59\% of the total variance, while the architecture contributes less than 7\%. We show the percentages with confidence intervals in \autoref{tab:variance_decomp}.



However, low ICC under \textsc{Corpus} variation does not underlie the same interpretation as low ICC under \textsc{Draw} or \textsc{Paraphrase}. A feature that scores differently on \texttt{Wikipedia} than on \texttt{Pile} may be legitimately domain-specific rather than unreliably scored. Despite these remarks, poor ICC persists even between \texttt{Pile} and \texttt{C4}, two general web text corpora, for which domain specificity cannot explain the instability. 
This instability within similar domains points to corpus sampling as a source of measurement noise rather than feature specialization.

\begin{figure}[H]
\centering
\includegraphics[width=0.7\linewidth, keepaspectratio]{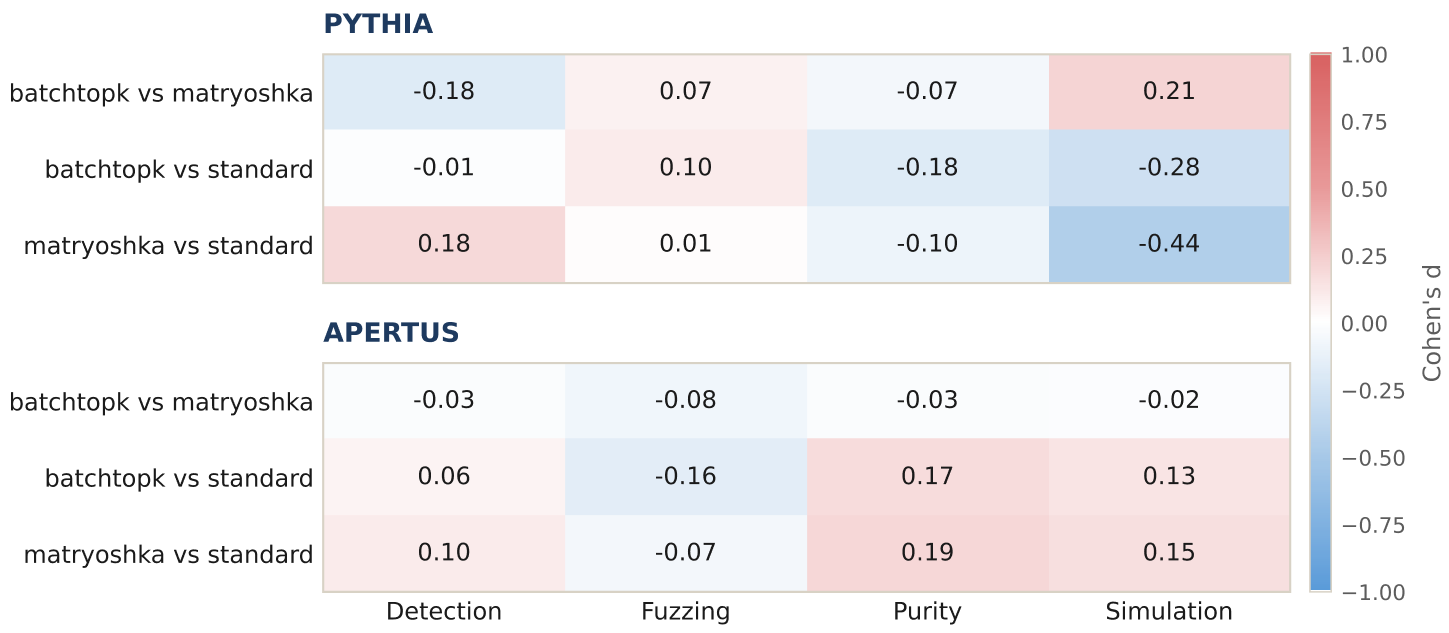}
\caption{Cohen's $d$ effect sizes between SAE architecture 
pairs on \texttt{Pythia-160M} (top) and \texttt{Apertus-8B} (bottom). We interpret Cohen's $d$ effect sizes as follows: $d = 0.2$ indicates a small effect, $d = 0.5$ a medium effect, and $d = 0.8$ a large effect \cite{sullivan2012effect}. Most 
effects are small to negligible and inconsistent in direction across metrics. }
\label{fig:mean_effect}
\end{figure}

To contextualize this directly, \autoref{fig:mean_effect} shows Cohen's $d$ effect sizes across three SAE architectures (\texttt{BatchTopK}, \texttt{Matryoshka}, \texttt{Standard}), with all methodological choices held fixed. Unlike within-architecture experiments, cross-architecture comparisons do not permit per-feature analysis via ICC or Jaccard, as features are not matched across architectures; therefore, we compare score distributions instead. On \texttt{Pythia-160M}, the largest pairwise effect is \texttt{Matryoshka} vs.\ \texttt{Standard} on \textsc{Simulation} ($d = -0.44$), falling in the small-to-medium range, yet the same pair shows $d = 0.01$ on \textsc{Fuzzing} and $d = -0.10$ on \textsc{Purity}. On \texttt{Apertus-8B}, all pairwise effects fall in the small range. Importantly, effect sizes are inconsistent in direction across metrics because a pair that scores positive and higher on one metric does not systematically score positive and higher on the others. The practical implication is direct: researchers comparing two SAE architectures using autointerpretability scores on a single corpus with a single random draw cannot reliably attribute observed differences in scores to architectural quality. 

\begin{figure}[H]
\centering
\includegraphics[width=0.7\linewidth, keepaspectratio]{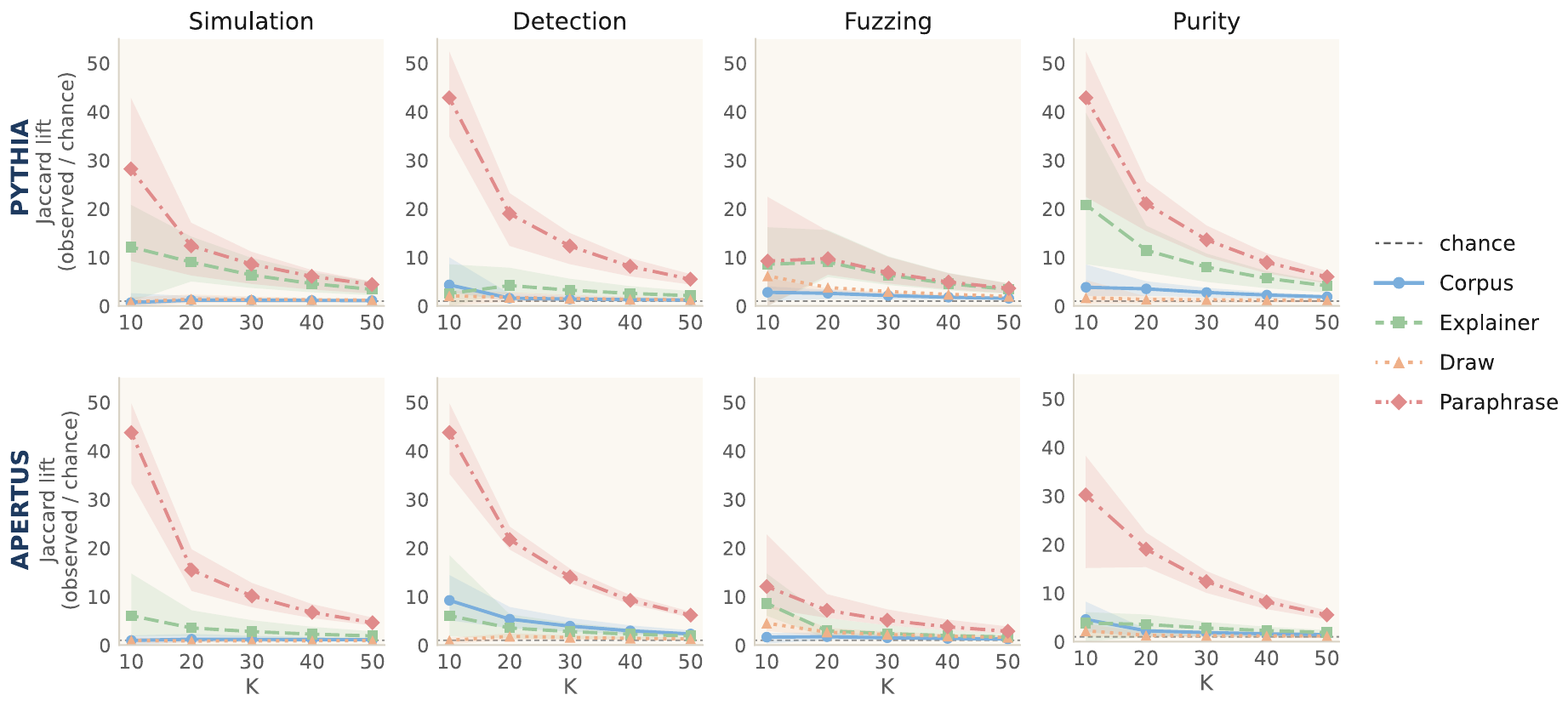}
\caption{The Jaccard lift for all four metrics across all conditions. \textsc{paraphrase}, which is the red line, shows the highest lift values across all conditions for all metrics.}

\label{fig:jaccard_lift}
\end{figure}

\subsection*{\texorpdfstring{\Itwo}{R2}~Each Metric has a Distinct Instability Profile}
\label{sec:findings_metrics}

\textit{\textsc{Simulation} and \textsc{Purity} are most sensitive to random draw; \textsc{Detection} is most sensitive to corpus choice; \textsc{Fuzzing} is unreliable across all sources. }

\begin{figure}
   \centering
\includegraphics[width=0.75\linewidth, keepaspectratio]{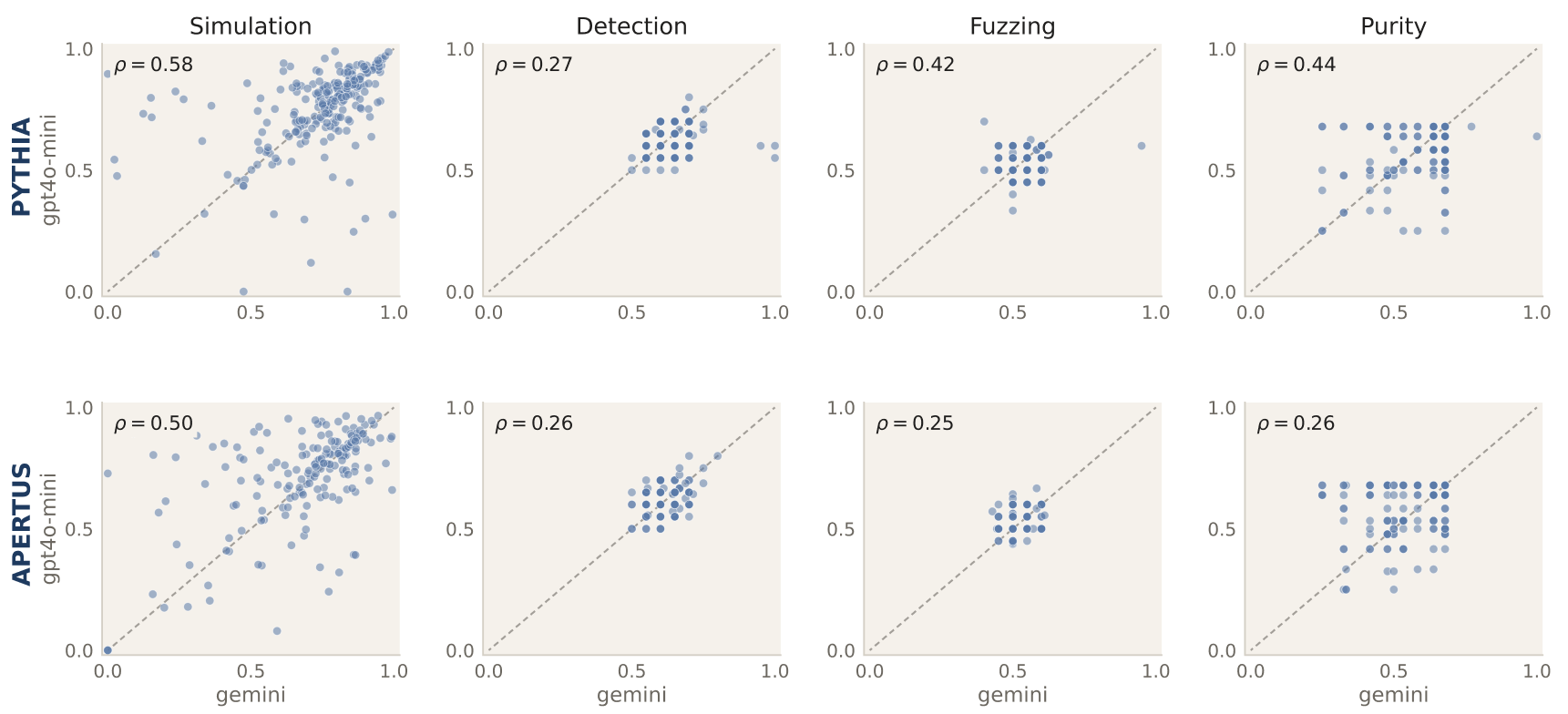}
\caption{Per-feature metric scores for \texttt{Gemini Flash} vs.\texttt{GPT4o-mini}; each point represents one feature.}
\label{fig:explainer_scatter}
\end{figure}

\paragraph{\textsc{Simulation}.}
\textsc{Simulation} shows high variance across conditions. It has a negative ICC value under the \textsc{Draw} variation, meaning repeated draws yield feature rankings that are \textit{less consistent than chance} (\autoref{tab:icc}). \textsc{Corpus} instability is similarly severe, with ICC near zero on \texttt{Pythia-160M} ($0.028$), yet \textsc{Simulation} becomes more reliable under \textsc{Explainer} variation; the absolute scores are volatile across explainers (\autoref{fig:explainer_scatter}). Under \textsc{Paraphrase}, ICC and Jaccard Lift (\autoref{tab:icc} and \autoref{fig:jaccard_lift}) are good for both models, confirming that \textsc{Simulation} responds to semantic content when evaluation inputs are held fixed.

\paragraph{\textsc{Detection}.}
\textsc{Detection} is the most stable metric overall, although still low ICC scores for \textsc{draw} and \textsc{Corpus}, though corpus sensitivity may partly reflect genuine feature domain-specificity rather than pure measurement noise. Under \textsc{Explainer} variation, ICC reaches $0.846\ [0.796, 0.886]$ on \texttt{Apertus-8B}. Researchers using different explainer models would obtain similar mean \textsc{Detection} scores but disagree substantially on which features are most interpretable (\autoref{fig:jaccard_lift}). Under \textsc{Paraphrase}, \textsc{Detection} shows the highest Jaccard lift of any metric ($42.87\times$ at $k=10$), confirming robustness to surface wording. 

\paragraph{\textsc{Fuzzing}.}
\textsc{Fuzzing} is unreliable across every variation, with the least reliable result for \textsc{corpus}. Despite achieving the most positive results among metrics on \textsc{draw}, the ICC and Jaccard Lift remain poor relative to other metrics across most conditions, even for \textsc{paraphrasing}. A metric that is not robust to paraphrasing responds to surface wording rather than semantic content. Combined with confidence intervals spanning zero for both \textsc{Corpus} and \textsc{Explainer}, we conclude that \textsc{Fuzzing} should not be used as a stand-alone reliability measure.

\paragraph{\textsc{Purity}.}
\textsc{Purity} measures the semantic coherence of activating examples, as reflected in the high variance for \textsc{Draw}, which accounts for 67\% and 62\% of its total variance across both models. This indicates that the semantic coherence of a feature is determined by \textit{which examples happen to be sampled}. Corpus sensitivity is partially expected: distributional shift across corpora may reflect genuine feature specialization or polysemanticity rather than pure measurement noise. Similar to \textsc{simulation} and \textsc{detection}, the highest ICC score is reached for \textsc{paraphrase}.

\subsection*{\texorpdfstring{\Ithree}{R3}~Feature Rankings show Instability beyond Mean Scores}
\label{sec:findings_rankings}

\textit{Mean scores appear stable across conditions, masking per-feature instability. Top-$k$ feature sets identified under one corpus or draw can have near-chance overlap with those identified under a different corpus or draw.}

\begin{figure}[t]
   \centering
   \includegraphics[width=0.65\linewidth, keepaspectratio]{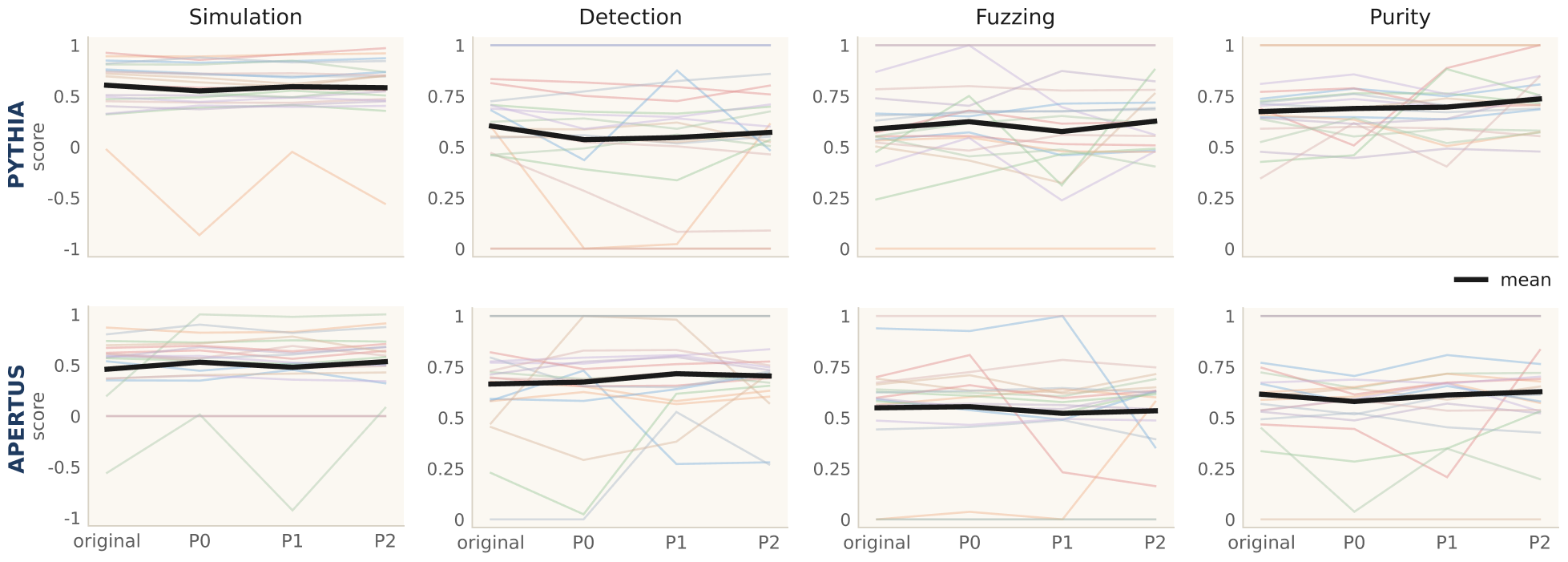}
\caption{Metric score trajectories of 20 features across three paraphrases. \textsc{Simulation} appears the least volatile, consistent with the ICC values in \autoref{tab:icc}.} 
\label{fig:trajectory_explainer_scatter}
\end{figure}

Mean scores are stable across most conditions (\autoref{tab:mean_scores}). For all four corpora, the mean \textsc{Simulation} score on \texttt{Pythia-160M} 
varies between 0.567 and 0.705, and mean \textsc{Detection} between 0.655 and 0.688. \autoref{fig:trajectory_explainer_scatter}, however, shows an example of how individual feature scores can be highly unstable. Trajectories cross frequently, and features that rank highly on one paraphrase do not reliably rank highly on another. The distinction matters because downstream applications, such as feature steering, circuit analysis, and manual inspection, often depend on selecting specific features rather than on reading population averages. \autoref{fig:jaccard_lift} and \autoref{tab:jaccard_lift} quantify ranking stability via Jaccard lift, but the interpretation differs by source of variation. We expand on this below. 


\paragraph{Draw, Explainer, and Paraphrase.}
For these three sources, lift should be high, since \textsc{Draw}, \textsc{Explainer}, and \textsc{Paraphrase} are arbitrary pipeline choices that should not affect which features are identified as interpretable. Low lift indicates that feature selection is a lottery rather than a stable property of the evaluation. Under \textsc{Draw} variation, \textsc{Simulation}, \textsc{Detection}, and \textsc{Purity} lift at $k = 10$ and $k = 20$ are near or below chance on both models, meaning that the ten or twenty highest-scoring features on one random draw have no reliable overlap with the top on another draw of the same corpus. \textsc{Paraphrase} consistently shows the highest lift of any source (\textsc{Simulation} and \textsc{Detection} reach $43.87\times$ at $k = 10$ on \texttt{Apertus-8B}), confirming that surface wording does not substantially reshuffle feature rankings for most metrics. \textsc{Explainer} lift is above chance for all metrics on both models, consistent with the ICC findings in \ref{sec:findings_metrics}. Interestingly, \textsc{Purity} has much higher lift on $k=10$ \texttt{Pythia-160M} (20.83) compared to \texttt{Apertus-8B} (3.82).

\paragraph{Corpus.}
The interpretation of corpus lift requires more care. Low lift under \textsc{Corpus} variation does not simply indicate unreliability; a feature that activates specifically on \texttt{Pile}-like text but not on \texttt{GitHub} code may be domain-specific rather than unreliably scored. Under this interpretation, corpus-dependent feature rankings reflect genuine properties of the features rather than measurement noise. However, domain specificity does not explain the instability observed across semantically similar corpora. \texttt{Pile} and \texttt{C4} are both general web text, yet yield a mean \textsc{Simulation} high mean delta (\ref{fig:delta_heatmap}) and below-chance Jaccard lift at $k = 10$ of $0.68\times$ on both models. This instability for domain-independent corpora is difficult to explain through feature specialization. It instead points to corpus sampling as a source of measurement noise that affects which activating examples are retrieved, and thereby which features appear interpretable.

\textsc{Corpus} variation also determines which features can be evaluated at all. The number of \texttt{Pythia-160M} features with at least one activating example ranges from 258 (\texttt{Pile}) to 215 (\texttt{Wikipedia}). When a hard requirement of exactly ten activating examples is applied, the number of features is reduced by 51\% on \texttt{Apertus-8B} (\autoref{tab:feature_coverage}). 

\section{Towards Reliable Evaluation}
\label{sec:recommendations}

We propose two complementary tools for the evaluation pipeline: a \textit{Stability Check} to assess the reliability of the evaluation before interpreting results, and a \textit{Minimum Reporting Checklist} to enable reproducible cross-paper comparisons. 

\subsection{Stability Check}
\label{sec:certificate}

We propose a \textit{Stability Check}, a test that researchers can run before interpreting autointerpretability scores as evidence of architectural quality, that goes beyond mean scores (\autoref{sec:findings_rankings}, \textbf{\textcolor{insightthree}{R3}}). The check quantifies the reliability of the evaluation by varying a methodological factor across $K$ conditions. We recommend varying \emph{random draws} as the default, as this is the lowest-cost source of replication and our results show it often shows the highest instability (\autoref{sec:findings_metrics}, \textbf{\textcolor{insightone}{R1}}, \textbf{\textcolor{insighttwo}{R2}}). A simpler check of semantic similarity does not work, as cosine similarity and BERTScore remain high even when scores are unstable \cite{zhang2020bertscore}, making direct stable measurements necessary (\autoref{app:similarity_values}).

\paragraph{How to run.}
For a chosen factor (such as \textsc{Draw}, \textsc{Corpus}, or \textsc{Explainer}), generate $K$ independent evaluation runs varying only that factor. Then 
compute:
\begin{itemize}
    \item \textbf{Mean ICC} across features: if ICC $< 0.50$, scores have poor reliability at the metric level. Results should be interpreted with caution and ideally replicated under additional conditions.
    \item \textbf{Top-$k$ Jaccard lift} across condition pairs: if lift is near $1\times$ (chance level), the top-$k$ feature set is unlikely to replicate under a different condition, and feature selection results have to be interpreted with caution.
\end{itemize}

\paragraph{How to interpret.}
\autoref{tab:certificate} summarises the thresholds. A metric that fails either threshold should not be used or used with caution to make cross-paper claims without methodological standardization. Passing both thresholds is a minimum condition for reliable comparison, not a guarantee of validity. ICC thresholds should be interpreted in context \cite{liljequist2019intraclass}: although $\text{ICC} \geq 0.75$ is the standard threshold for good reliability, note that lower values can be acceptable depending on the application and the broader pattern of reported ICC values across conditions. Not all metrics fail in the same way (\textbf{\textcolor{insighttwo}{R2}}). Moreover, the variance decomposition (\autoref{par:variance_decomp}) can be applied if multiple methodological factors are varied simultaneously, but at a higher computational cost scaling with $N_{\text{setup}} \times N_{\text{conditions}} \times N_{\text{features}}$.

\begin{table}
    \centering
    \caption{\textbf{Stability Check thresholds.} ICC interpretation follows 
    \cite{koo2016guideline}; Jaccard lift threshold of $2\times$ indicates 
    feature selection above chance \cite{kuncheva2007stability}. }
    \label{tab:certificate}
    \footnotesize 
    \begin{tabular}{llcc}
        \toprule
        \textbf{Level} & \textbf{Measure} & \textbf{Pass} 
        & \textbf{Interpretation if failed} \\
        \midrule
        Metric  & Mean ICC & $\geq 0.5$ 
        & Scores unreliable across conditions \\
        Feature & Top-$k$ Jaccard lift & $\geq 2\times$ 
        & Feature selection at or near chance \\
        \bottomrule
    \end{tabular}
\end{table}

\subsection{Minimum Reporting Checklist}
\label{sec:checklist}

To enable cross-paper comparison, we propose a \textit{Minimum Reporting Checklist} for autointerpretability evaluation (\autoref{tab:checklist}). Papers that omit these details cannot be reliably compared to prior work, as our findings demonstrate that each omitted factor is a potential source of substantial score variance (\textbf{\textcolor{insightone}{R1}}).

\begin{table}[t]
    \centering
    \caption{\textbf{Minimum Reporting Checklist}, where each field corresponds to a source of variation identified in our 
    experiments.}
    \label{tab:checklist}
    \footnotesize 
    \begin{tabular}{lp{10cm}}
        \toprule
        \textbf{Field} & \textbf{What to report} \\
        \midrule
        \textsc{Corpus}   & Name, size, and domain characteristics \\
        \textsc{Sampling} & Number of activating examples and selection method (e.g., top-10 from a pool of 100) \\
        \textsc{Draws}    & Number of independent draws; if only one draw is used, acknowledge as a limitation \\
        \textsc{Explainer} & Model name and version \\
        \textsc{Stability} & At least one stability metric: ICC and/or top-$k$ Jaccard lift \\
        \bottomrule
    \end{tabular}
\end{table}

Each field in the checklist corresponds directly to a source of variation studied in this paper. Reporting the \textsc{Corpus} and \textsc{Draws} addresses 
the two most harmful sources of instability identified in our experiments (\textbf{\textcolor{insighttwo}{R2}}, \textbf{\textcolor{insightthree}{R3}}). Reporting the \textsc{Explainer} enables replication across experiments using different models. Reporting at least one stability metric, along with the evaluation settings and results, provides a more complete picture for reproducible evaluation. 

\section{Conclusion}
\label{sec:conclusion}

Autointerpretability scores are the dominant means of assessing SAE feature interpretability, and cross-paper comparisons of SAE architectures rely on the reliability of these scores. We asked whether \textit{current autointerpretability evaluation practices are reliable enough to support cross-paper comparisons of SAEs, or do pipeline choices determine the outcome?} To answer this, we systematically varied four axes of the evaluation pipeline, across two models (\texttt{Pythia-160M} and \texttt{Apertus-8B}) and four metrics, measuring instability via ICC, Top-$k$ Jaccard lift, and REML-based variance decomposition.

Our findings are summarised in threefold. \textbf{\textcolor{insightone}{(R1)}}~Methodological variance collectively exceeds architectural variance in all metrics and models, with pipeline choices accounting for more score variance than SAE architectures. \textbf{\textcolor{insighttwo}{(R2)}}~Each metric has a distinct instability profile, with \textsc{Detection} being most stable overall, while \textsc{Fuzzing} is unreliable across all sources of variation. \textbf{\textcolor{insightthree}{(R3)}}~Mean scores mask per-feature instability, as top-$k$ feature rankings have near-chance overlap across corpus and draw conditions, with consequences for applications that depend on selecting specific features. Additional results show that explanation similarity cannot diagnose score instability, as similarity scores remain high even when autointerpretability scores vary, ruling out this cheap diagnostic.

Several limitations bound the scope of our conclusions. Our experiments use a fixed scorer model (\texttt{Gemini Flash}) and a fixed layer (layer 8 for \texttt{Pythia-160M}, layer 16 for \texttt{Apertus-8B}); scorer variation and layer-dependence are additional axes of instability we did not study and that can influence scores \cite{puri2025fade}. Feature samples are limited to 100--267 per condition, and variance estimates from the decomposition should be interpreted as approximate, given the partially crossed experimental design. Additionally, including more metrics can provide a more robust evaluation \cite{paulo2025automatically, puri2025fade}. Finally, different numbers of corpora, draws, explainers, and paraphrases could have been tested to provide a richer picture of the influence of these choices. Under the specific pipeline configurations tested, the methodological variance component exceeds the architectural variance component. We do not claim this generalizes to all scorers, layers, or SAE architectures.

These findings have a direct bearing on the ongoing research on SAE utility \cite{kantamneni2025are}. 
To support more reliable evaluation, we provide a \textit{Stability Check} and a \textit{Minimum Reporting Checklist} (\autoref{sec:recommendations}), allowing researchers to diagnose evaluation reliability before interpreting score differences as architectural effects, and to report enough methodological detail to make cross-paper comparison reproducible.

Future work could examine alternative evaluations that do not rely solely on LLM-judged explanations. Causal and interventionist evaluations \cite{makelov2025towards, marks2025sparse} validate features through their downstream effects on model behavior. Similar lines of work use SAE features for steering interventions \cite{chalnev2024improving}, whose effects can be measured directly on model outputs rather than mediated by an explanation step. On the other hand, the discussion about the usefulness of SAEs \cite{li2026evaluating, kantamneni2025are, wu2025axbench} underscores the need for evaluations with established reliability. We view causal, interventionist, and explanation-based evaluations as complementary: explanation-based scoring scales to millions of features in ways that causal methods currently do not, while interventionist evaluations provide the behavioral grounding that explanation quality alone cannot establish. Establishing the reliability of each approach and identifying the conditions under which they agree are important directions for future work.

As we see it, the goal is not to abandon autointerpretability, as it currently remains the most scalable approach to feature-level interpretation. The goal is to use it with the caution its instability warrants, and to build the shared reporting standards that would allow the field to accumulate reliable evidence.

\bibliography{main}

\clearpage
\appendix

\section{Related Work}
\label{app:related_work}

\paragraph{Evaluation Gap}
The field has recognized that improvements in the sparsity-reconstruction trade-off often do not directly translate to improved monosemanticity \cite{minegishi2025rethinking, karvonen2025saebench}. The intuition behind reconstruction error as a proxy is that a low error indicates the SAE faithfully captures the original model's activations \cite{bricken2023monosemanticity, rajamanoharan2024improving}. Similarly, sparsity is assumed to promote interpretability by forcing the model to represent each input using few active latent units \cite{cunningham2024sparse}. However, SAEs can be optimized along this trade-off and still produce polysemantic, unstable, or irrelevant features \cite{karvonen2025saebench, minegishi2025rethinking}. More broadly, most metrics used in practice are \textit{proxy} metrics: reconstruction error, sparsity, and loss recovery reflect optimization behavior rather than semantic quality, and optimizing for sparsity can introduce failure modes such as feature absorption \cite{chanin2024absorption} and dead latents \cite{gao2025scaling}.

A first step toward a more principled approach was taken by \cite{makelov2025towards}, who proposed a framework of faithfulness checks and feature-editing experiments relative to a supervised baseline. The authors explicitly acknowledge that their framework depends on a particular parametrization of task-relevant variables, which makes comparisons across settings difficult. This is a limitation that reflects the broader challenge that interpretability evaluation is conditional on task, variables, and the conceptual lens. SAEBench \cite{karvonen2025saebench} extends this toward standardization, comparing SAE variants across reconstruction fidelity, feature disentanglement, interpretability, and concept detection. They show that different model choices, such as dictionary size, influence evaluation metrics. A key finding is that these metrics cannot be unified into a single score, making evaluation fundamentally multi-objective: metric choice depends on the researcher's priority.

\paragraph{Autointerpretability}
Interpretability evaluation in practice relies predominantly on \textit{automated interpretability} pipelines: a LM generates natural language explanations for features, and a scorer evaluates those explanations by testing whether they predict feature activations in new contexts. The approach was originally developed for neuron-level interpretation \cite{bills2023language} and subsequently adapted to SAE features \cite{bricken2023monosemanticity, cunningham2024sparse}. Recent work has extended the pipeline in several directions. \cite{paulo2025automatically} scale explanation generation to millions of features, while \cite{puri2025fade} examines failure modes of automated description. Additionally, \cite{puri2025fade} shows that explanations may not faithfully capture feature behavior and that scoring procedures can reward misleading descriptions.

\paragraph{Known Sources of Variation}
Several studies have identified factors that influence autointerpretability scores, motivating the systematic analysis we present in this paper. We organize these into three groups.

\begin{itemize}
    \item  \textit{SAE and model properties.} Interpreting larger SAEs is more difficult, with larger dictionaries yielding lower scores \cite{puri2025fade}.  Scores vary across layers, with later layers tending to be more polysemantic and harder to explain \cite{puri2025fade}. SAEs with more features tend to score higher than individual neurons, suggesting that sparsity aids interpretability \cite{paulo2025automatically}. Score variation on detection across SAE architectures is documented in \cite{karvonen2025saebench}.

\item \textit{Example selection.} The choice of which activating examples are shown to the explainer affects explanation quality. Using only top-activating examples yields concise, specific descriptions but fails to capture the full activation distribution \cite{paulo2025automatically}. Random sampling yields interpretations that span a more diverse set of activating contexts, whereas stratified sampling can produce descriptions that are overly broad \cite{paulo2025automatically}. Including only activating contexts increases specificity but reduces sensitivity; including non-activating contexts has the opposite effect \cite{paulo2025automatically}. Providing a larger number of examples yields slightly higher scores overall \cite{paulo2025automatically, puri2025fade}, though this improvement may reflect increased generality rather than increased accuracy \cite{paulo2025automatically}. 

\item \textit{Explainer and scorer properties.} Larger explainer models produce better 
feature descriptions and higher scores \cite{paulo2025automatically, 
puri2025fade}. For simpler metrics such as detection and fuzzing, using a smaller scorer model leads to lower scores, suggesting that scorer capacity interacts with metric complexity \cite{paulo2025automatically}. Prompt construction affects results \cite{puri2025fade}, as does the use of few-shot prompting \cite{puri2025fade}. Notably, \cite{paulo2025automatically} finds that training dataset variation has little effect on scores, suggesting that corpus choice at training time is less influential than corpus choice at evaluation time, which is a distinction our experiments make explicit.
\end{itemize}

What remains absent from prior work is a systematic quantification of the 
\textit{relative} contribution of these sources to total score variance, and an 
assessment of whether their combined effect exceeds the variance attributable to 
architectural differences. This paper addresses both questions, with additional focus on corpus, draws, explainers, phrasing, and architecture.  

\section{Experiments}
\label{app:experiments}
The following sections contain additional information to help with reproducibility of the experiments. 

\subsection{Experiment Overview}
An overview of the number of variations for each condition can be found in \autoref{tab:experiments}. 

\paragraph{Exp. 1 — Corpus dependence.} We generated independent explanations and scores for the same set of features on four different corpora spanning a gradient of distributional distance from the training data: the Pile (in-distribution), C4 (similar), Wikipedia (moderate), and GitHub code (dissimilar). All datsets are listed in \autoref{tab:corpora}. This tests whether autointerpretability scores reflect stable feature properties or are confounded by the choice of evaluation corpus. It is expected that more similar corpora have similar scores on the same features. 

\paragraph{Exp. 2a — Sampling instability.} For a fixed corpus, we generated multiple explanations per feature with the same explainer by drawing different random subsets of activating examples. This isolates the variance introduced by which examples happen to be sampled, holding all other factors constant. We use five independent draws per feature to balance statistical and computational considerations.

\paragraph{Exp. 2b — Explainer model dependence.} We explained the same features using two different explainer LLMs: \texttt{Gemini Flash} and \texttt{GPT4o-mini}. All explanations were scored with a fixed, but different model. Both are small, efficient models positioned below the frontier tier of their respective families, and are broadly comparable in general capability. As other papers have looked into the effect of the size of the explainers (\cite{paulo2025automatically, puri2025fade}, this is not the main focus. As the chosen models are comparable in size, we investigated the effects of their differences, which should be minimal in theory. We restricted our comparison to two closed-source models representing the most common practical choice; whether instability patterns generalize to open-source explainers remains an open question.

\paragraph{Exp. 2c — Paraphrase sensitivity.} To test whether scoring measures semantics or primarily surface wording, we generated three paraphrases of each explanation and scored all versions on the same held-out sequences. If scores vary substantially across paraphrases, this indicates that the scorer is responding to wording rather than meaning. We generate three paraphrases per explanation, again balancing statistical and computational considerations. Critically, paraphrase represents the most semantically conservative perturbation in our experimental design, as it focuses solely on the explanation itself, not on the input to or generation of the explanation.  

\paragraph{Exp. 3 — Cross-architecture comparison.} To contextualize the methodological instability, we compared score distributions across three SAE architectures trained on the same model and data: a standard ReLU SAE, a Matryoshka SAE, and a BatchTopK SAE. All other methodological choices were fixed. This provides a reference point: if methodological variance (Exp. 1–2) exceeds architectural variance, this could limit the validity of cross-SAE comparisons in the literature. We investigate SAEs trained on smaller and larger models (Pythia160M vs. Apertus-8b Instruct). Earlier work has shown that architectures can have a different autointerpretability score \cite{karvonen2025saebench}, but we extend this to multiple metrics.  

\begin{table}[t]
\centering
\caption{Experimental design. Each experiment isolates one source of variation.}
\label{tab:experiments}
\small
\begin{tabular}{llc}
\toprule
\textbf{Experiment} & \textbf{Factor Varied} & \textbf{Variation} \\
\midrule
Corpus & Evaluation corpus & 4 corpora \\
\midrule
Sampling & Random draw of examples & 5 draws \\
Explainer & Explainer LLM &2 explainer models \\
Phrasing & Explanation wording& 3 paraphrases \\
\midrule
Architecture & Architecture influence & 3 architectures \\
\bottomrule
\end{tabular}
\end{table}

\subsection{Corpora}
\label{app:corpora}
All corpora are sampled to 10,000 sequences, tokenized with the model-specific tokenizer (Pythia tokenizer for Pythia-160M; Apertus tokenizer for Apertus-8B), and truncated to 32 tokens following the findings of \cite{paulo2025automatically}. All datasets were accessed through HuggingFace. \autoref{tab:corpora} summarises corpus statistics.

The four corpora were chosen to span different degrees of similarity to the SAE training distribution, from similar to dissimilar. Both SAEs are trained on the Pile. 

\begin{table}[H]
\centering
\caption{Corpus statistics. All corpora sampled to 10,000 sequences of 32 tokens.}
\label{tab:corpora}
\small
\begin{tabular}{lllll}
\toprule
\textbf{Corpus} & \textbf{Domain} & \textbf{HuggingFace identifier} 
& \textbf{License} & \textbf{Experiments} \\
\midrule
The Pile      & General web text (mixed) 
  & \texttt{monology/pile-uncopyrighted} & Other & All \\
C4            & Cleaned web text         
  & \texttt{allenai/c4} & ODC-BY & Exp.\ 1 \\
Wikipedia     & Encyclopedic text       
  & \texttt{wikimedia/wikipedia} & CC-BY-SA-3.0 & Exp.\ 1 \\
GitHub Code   & Source code              
  & \texttt{codeparrot/github-code} & Other & Exp.\ 1 \\
\bottomrule
\end{tabular}
\end{table}

\subsection{Feature Sampling}
\label{app:feature_sampling}
\begin{figure}[H]
\includegraphics[width=1.0\textwidth]{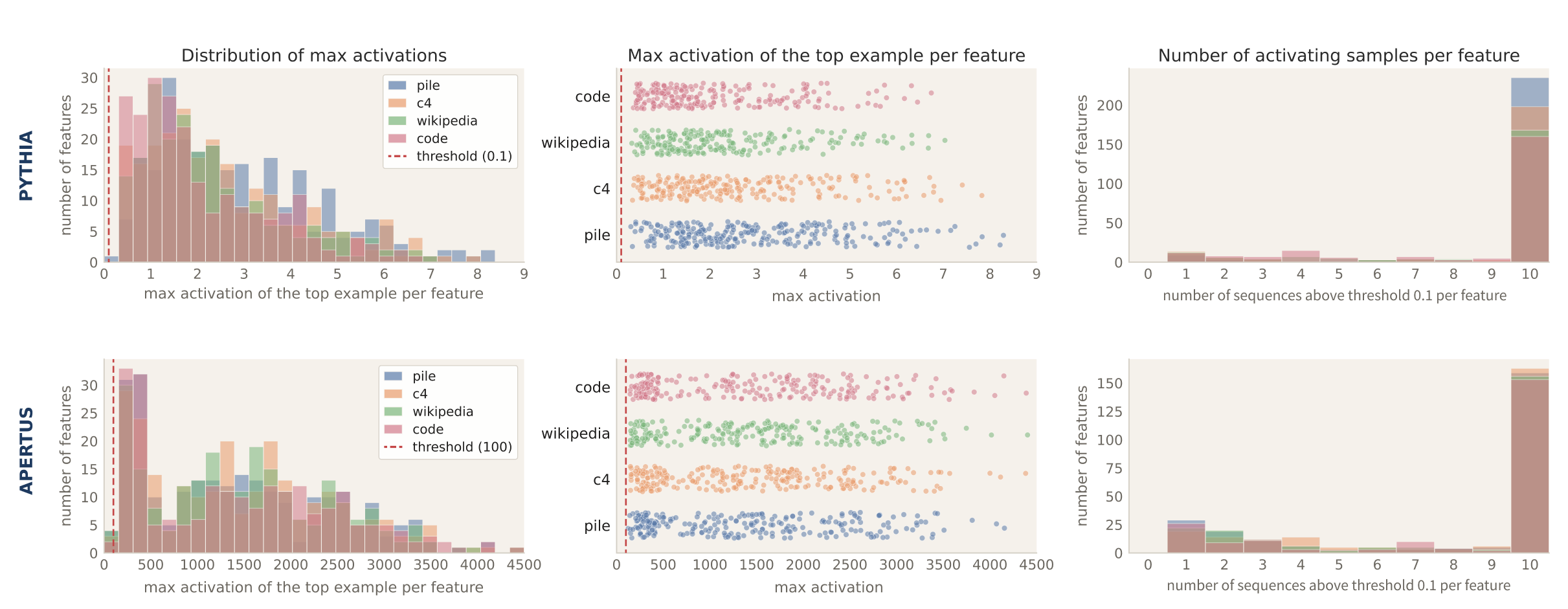}
\caption{Activation diagnostics for features, considering all four datasets. The first column of figures shows the distribution of the highest-activating sample for each feature. The second shows the spread of features between the minimum activation threshold and the maximum activation detected for a feature. The final one shows how many activating samples there are per feature.}
\label{fig:activation_diagnostic}
\centering
\end{figure}

Features are selected at random from the full SAE feature pool per experiment, targeting to retrieve 250 active features per condition. To reach this number, we randomly sample 300 features for Pythia and 2000 features for Apertus. A feature is considered \emph{active} if its maximum activation over the evaluation corpus exceeds a model-specific threshold: 0.1 for Pythia-160M and 100 for Apertus-8B, reflecting the difference in activation scales between the two models. Figure \ref{fig:activation_diagnostic} shows that these thresholds yield sufficient activating sequences. 

For each active feature, the 10 highest-activating sentences are extracted for explanation generation. Detection scoring uses 10 randomly sampled negative sentences (where the feature does not activate) per feature. We have opted for the standard 10 highest-activating sequences, as prior work has already shown their influence, and we have chosen a number that is feasible regarding computing costs and efficiency \cite{paulo2025automatically, puri2025fade}. However, the requirement of at least one activating sample per feature can yield explanations with fewer than 10 samples. An example of this is shown in \autoref{tab:feature_coverage} for the \textsc{corpus} condition.

\begin{table}[H]
\centering
\caption{Number of features \textbf{with at least one} activating example per corpus vs. \textbf{with 10 activating samples per corpus}. \textit{Overlap} denotes features activating in all four corpora simultaneously and constitutes the evaluation set for Exp.~1. \textit{Distinct} indicates the number of different features in total}
\label{tab:feature_coverage}
\small
\begin{tabular}{llrrrrrrr}
\toprule
& & \multicolumn{5}{c}{\textbf{Corpus}} & & \\
\cmidrule(lr){3-7}
\textbf{Model} & \textbf{Threshold} 
  & \textbf{Pile} & \textbf{C4} & \textbf{Wiki.} & \textbf{Code} 
  & \textbf{Overlap} & \textbf{Distinct} \\
\midrule
\multirow{2}{*}{Pythia-160M} 
  & $\geq 1$  & 258 & 236 & 215 & 223 & 190 & 258 \\
  & 10 & 143 & 134 & 131 & 133 & 121 & 143 \\
\midrule
\multirow{2}{*}{Apertus-8B}  
  & $\geq 1$  & 244 & 244 & 234 & 221 & 189 & 264 \\
  & 10 & 130 & 130 & 128 & 128 & 128 & 130 \\
\bottomrule
\end{tabular}
\end{table}

\subsubsection{Feature Evaluation and Activation Coverage}
\label{app:feature_eval}
There are differences in the number of features that can be evaluated per methodological condition, which is shown in for the \textsc{Corpus} condition in \autoref{tab:feature_coverage}. This also introduces a small selection bias, as the choice of corpus determines which features can be scored. 

\subsubsection{Draw Sampling (Exp. 2a)}
\label{app:draw_sampling}
For Exp. 2b, each of the five draws samples 10 sequences uniformly at random from a pool of 100 top-activating sequences, yielding an independent explanation and score profile for each draw. This experiment is close to the experiments on top-activating and stratified sampling reported in \cite{paulo2025automatically}.

\subsection{Explainer Prompt}
\label{app:explainer_prompt}

\begin{promptbox}{System prompt — Explanation generation}
You are a neuron interpreter for a sparse autoencoder (SAE) 
trained on a language model.
Your task: Given examples of text where a specific SAE feature 
activates strongly, describe what concept, pattern, or behavior 
the feature detects.
Guidelines:
- Be specific and concise (1-2 sentences)
- Focus on the common pattern across all examples
- Mention token-level patterns if relevant
- If the pattern is unclear, say so honestly
- Describe the concept directly (e.g. "Mathematical equations 
  and formulas" or "Python source code and programming 
  constructs"). Do NOT start with "This feature activates 
  for/when/on" or similar phrases.
\end{promptbox}

\vspace{4pt}

\begin{promptbox}{User prompt — Explanation generation}
Here are text examples where SAE feature {feature_id} activates 
strongly. Each example shows the text and the activation 
strength.
{examples}
Based on these examples, describe what concept or pattern 
feature {feature_id} detects. {word_limit_instruction}starting 
directly with the concept (not with "This feature 
activates...").
\end{promptbox}

\vspace{4pt}
\noindent where \texttt{\{word\_limit\_instruction\}} resolves to:
\begin{itemize}
    \item \textit{Default:} ``Answer in 1--2 sentences, ''
    \item \textit{W10 condition:} ``Answer in at most 10 words, ''
\end{itemize}

\subsection{Explainers (Exp. 2b)}
\label{app:explainers}

Exp. 2b compares two explainer LLMs: \texttt{Gemini Flash} (primary) and \texttt{GPT4o-mini}(comparison). Both models receive identical inputs, which are the same ten highest-activating sequences with activating tokens highlighted, and are prompted to generate a natural language description of the feature. All scoring in Exp. 2b uses \texttt{Gemini Flash} as the fixed scorer, isolating the effect of explainer model identity 
from scorer variation.

\subsection{Paraphrase Generation (Exp. 2c)}
\label{app:paraphrase}

Three paraphrases are generated per explanation. The paraphrase prompt instructs the model to preserve the semantic meaning of the original explanation while varying the surface wording, sentence structure and vocabulary. Each original explanation and its three paraphrases are scored indepdently on the same held-out sequences, producing four observations per feature per metric. 

Paraphrases are generated by the same LLM as the scorer, which is Gemini Flash. This could mean that our paraphrase stability result may overstate semantic invariance because both the paraphraser and scorer share representational biases. A future test would use paraphrases from a structurally different model, which we view this as an important follow-up.

\begin{promptbox}{System prompt}
You are a precise language model assistant. Your task is to 
rephrase a given feature description in multiple ways, preserving 
the exact same meaning each time. Vary only vocabulary and sentence 
structure - do NOT change the concept described. Return ONLY a JSON 
array of strings, one per paraphrase. No extra text.
\end{promptbox}

\vspace{4pt}

\begin{promptbox}{User prompt}
Original description: {explanation}
Generate {n} paraphrases of the above description. Keep the exact 
same meaning. Return ONLY a JSON array of {n} strings.
\end{promptbox}

\subsection{Explanation Similarity}
\label{app:explanation_similarity_method}
To measure semantic consistency of explanations across conditions, we compute two similarity measures:

\begin{itemize}
    \item \textbf{Cosine similarity} of E5-small-v2 embeddings (\texttt{intfloat/e5-small-v2}). Each explanation is embedded using mean pooling over token embeddings, and pairwise cosine similarity is computed across all condition pairs for each feature.
    \item \textbf{BERTScore O1} \citep{zhang2020bertscore} using RoBERTa-large as the reference model. BERTScore captures token-level lexical overlap weighted by contextual embeddings, 
    providing a complementary measure less sensitive to synonym substitution than cosine similarity.
\end{itemize}

Both measures are all reported as mean pairwise within-experiment comparisons. The two measures are highly consistent across all experiments and models (\autoref{tab:similarity}).

\subsection{Scoring Procedure}
\label{app:scoring}
In the following section, we extend with details on the metrics used and the scoring prompts. 

\subsubsection{Metric Details}
\label{app:metric_details}
The first three metrics follow a two-phase pipeline: 1) the explainer LLM generates a natural language description of the feature from its top-activating samples and 2) a scorer LLM evaluates the explanation against held-out sequences. The last metric, \textsc{Purity}, does not depend on any LLM. 

\begin{itemize}
    \item \textsc{Simulation} \citep{bills2023language}. A scorer LLM is shown the explanation and a set of sequences and is asked to predict the feature's activation at each token position on a scale of 1 to 10. The simulation score is the Pearson correlation between predicted and actual activation values at the token level, ranging from -1 (inverse prediction) to 1 (perfect prediction). A score of 0 indicates that predictions are no better than chance. 
    \item \textsc{Detection} \citep{paulo2025automatically}. The scorer LLM is shown the explanation and is asked to classify sequences as either activating or non-activating. The detection score is the AUC of a binary classifier, ranging from 0 to 1, with 0.5 indicating chance performance. Positive sequences are the top 10 activating sequences; negative sequences are 10 randomly sampled sequences from the corpus. Detection is "cheaper" compared to simulation, as it requires only a sequence-level binary decision, rather than token-level activation prediction/  classification of activating and non-activating samples. 
    \item \textsc{Fuzzing} \citep{paulo2025automatically}.  Similar to detection, but then on the token level within activating sequences. The scorer is shown an explanation and sequences with some tokens masked, and asked to predict whether the original token is likely to activate the feature. The fuzzing score is the AUC of these token-level predictions, ranging from 0 to 1. Fuzzing is more sensitive to whether the explanation correctly identifies which tokens are relevant. 
    \item \textsc{Purity} \citep{puri2025fade}. Measures whether high activations are exclusive to the target concept at the sequence level. Natural samples are rated by a scorer LLM on a three-point scale (0 = not expressed, 1 = partial, 2 = clearly expressed); ambiguous cases are discarded, leaving binary concept/non-concept labels. Average Precision (AP) is then computed using feature activation magnitudes as the ranking signal, ranging from 0 to 1. A high Purity score indicates that strongly activating sequences are specific to the target concept, while a low score indicates that unrelated sequences also produce high activations. 
\end{itemize}

\subsubsection{Scoring Prompts}
\label{app:scorer_prompts}

We use four scoring prompts corresponding to the four metrics. All prompts are sent to \texttt{Gemini Flash} as the scorer. The \texttt{\{explanation\}} placeholder is replaced with the 
generated feature explanation; other placeholders are filled 
with the relevant sequences.

\paragraph{Simulation}
Implements the scoring procedure of \cite{bills2023language}. 
The scorer predicts token-level activation magnitudes on a 
0--10 scale; the simulation score is the Pearson correlation 
between predicted and actual activations.

\begin{promptbox}{System prompt — Simulation}
You are a neuron activation simulator for a sparse autoencoder 
(SAE) trained on a language model.

Given a description of what an SAE feature detects and multiple 
pieces of text, predict how strongly the feature would activate 
on each text.

Return a JSON array of numbers between 0 and 10:
- 0 means the feature would not activate at all
- 10 means the feature would activate very strongly

Return ONLY the JSON array, nothing else. 
Example: [3.5, 0.0, 8.2, 1.0]
\end{promptbox}

\vspace{4pt}

\begin{promptbox}{User prompt — Simulation}
Feature description: {explanation}

Texts:
{texts}

For each text, predict the feature activation (0-10). 
Return ONLY a JSON array of {num_texts} numbers.
\end{promptbox}

\vspace{8pt}

\paragraph{Detection}
Implements the detection procedure of \cite{paulo2025automatically}. 
The scorer performs binary classification of sequences as 
activating (1) or non-activating (0); the detection score is 
the AUC over positive and negative sequences.

\begin{promptbox}{System prompt — Detection}
You are an intelligent and meticulous linguistics researcher.
You will be given a certain feature of text, such as "male 
pronouns" or "text with negative sentiment". You will then be 
given several text examples. Your task is to determine which 
examples possess the feature.
For each example in turn, return 1 if the sentence is correctly 
labeled or 0 if the tokens are mislabeled. You must return your 
response in a valid Python list. Do not return anything else 
besides a Python list.
\end{promptbox}

\vspace{4pt}

\begin{promptbox}{User prompt — Detection}
feature interpretation: {explanation}
Text examples:
{texts}
\end{promptbox}

\vspace{8pt}

\paragraph{Fuzzing}
Implements the token-level fuzzing procedure of 
\cite{paulo2025automatically}. Activating tokens are marked 
with \texttt{<<} and \texttt{>>} delimiters; the scorer 
classifies whether the marked tokens correctly represent the 
feature. The fuzzing score is the AUC over correctly and 
incorrectly marked token spans.

\begin{promptbox}{System prompt — Fuzzing}
You are an intelligent and meticulous linguistics researcher.
You will be given a certain feature of text, such as "male 
pronouns" or "text with negative sentiment". You will be given 
a few examples of text that contain this feature. Portions of 
the sentence which strongly represent this feature are between 
tokens << and >>.
Some examples might be mislabeled. Your task is to determine 
if every single token within << and >> is correctly labeled. 
Consider that all provided examples could be correct, none of 
the examples could be correct, or a mix. An example is only 
correct if every marked token is representative of the feature.
For each example in turn, return 1 if the sentence is correctly 
labeled or 0 if the tokens are mislabeled. You must return your 
response in a valid Python list. Do not return anything else 
besides a Python list.
\end{promptbox}

\vspace{4pt}

\begin{promptbox}{User prompt — Fuzzing}
feature interpretation: {explanation}
Text examples:
{examples}
\end{promptbox}

\vspace{8pt}

\paragraph{Purity}
Implements the FADE sequence rating procedure of 
\cite{puri2025fade}. The scorer rates each sequence on a 
0--2 scale according to how clearly the concept is expressed; 
the purity score is the average precision over rated sequences.

\begin{promptbox}{System prompt — Purity}
You are tasked with building a database of sequences that best 
represent a specific concept.
To create this, you will review a dataset of varying sequences 
and rate each one according to how much the concept is 
expressed.
For each sequence, assign a rating based on this scale:
0: The concept is not expressed.
1: The concept is vaguely or partially expressed.
2: The concept is clearly and unambiguously present.
Use conservative ratings. If uncertain, choose a lower rating 
to avoid including irrelevant sequences in your database.
If no sequence expresses the concept, rate all sequences as 0.
Each sequence is identified by a unique ID. Provide your 
ratings as a Python dictionary with sequence IDs as keys and 
their ratings as values.
Example output: {"14": 0, "15": 2, "20": 1, "27": 0}
Output only the dictionary - no additional text, comments, 
or symbols.
\end{promptbox}

\vspace{4pt}

\begin{promptbox}{User prompt — Purity}
Concept: {explanation}

Sequences:
{sequences}
\end{promptbox}

\vspace{8pt}

\subsubsection{Paraphrase Generation Prompt}
\label{app:paraphrase_prompt}

Paraphrases are generated with the following prompts. The system prompt instructs the model to preserve semantic meaning while varying surface form; the user prompt specifies the original explanation and the number of 
paraphrases required ($n = 3$). Responses are parsed from the returned JSON array.

\begin{promptbox}{System prompt — Paraphrase generation}
You are a precise language model assistant. Your task is to 
rephrase a given feature description in multiple ways, 
preserving the exact same meaning each time. Vary only 
vocabulary and sentence structure - do NOT change the concept 
described. Return ONLY a JSON array of strings, one per 
paraphrase. No extra text.
\end{promptbox}

\vspace{4pt}

\begin{promptbox}{User prompt — Paraphrase generation}
Original description: {explanation}
Generate {n} paraphrases of the above description. Keep the 
exact same meaning. Return ONLY a JSON array of {n} strings.
\end{promptbox}

\section{Instability Quantification}
\label{app:instability_quant}

\subsection{ICC specification}
\label{app:icc_spec}
We use the Intraclass Correlation Coefficient (ICC) to quantify the proportion of total score variance attributable to stable \textbf{between-feature} difference, rather than within feature noise \cite{koo2016guideline}. 

There is a difference between ICC(1,1) and ICC(2,1):

\begin{itemize}
    \item \textbf{Draw — ICC(1,1).} Random draws of activating examples are not a fixed, replicable set of conditions. Each draw is a unique random sample from the activation pool. We therefore treat draws as a one-way random effects model (ICC(1,1)), in which the conditions themselves are sampled from a larger population of possible draws.

    \item \textbf{Corpus, Explainer, Paraphrase — ICC(2,1).} These sources involve a fixed, replicable set of conditions (four specific corpora; two specific explainer models; four specific explanation versions). We therefore treat them as a two-way mixed effects model (ICC(2,1)), in which conditions are fixed and features are random.
\end{itemize}

However, even though these thresholds are stated, this does not indicate that anything with less than 0.75 is directly unreliable \cite{liljequist2019intraclass}. It can be informative to look at other studies that report ICC, to receive an indication of the range of ICC values. We report the ICC for each metric (simulation, detection, fuzzing, purity) and for each source of variation (corpus, draw, explainer).

\subsection{Top-\textit{k} Jaccard Lift}
\label{app:jaccard_lift}

Raw Jaccard values are not directly comparable across experiments because the feature pool size $N$ differs between experiments. We therefore normalise by the expected Jaccard under random selection:
\begin{equation}
    J_k^{\text{chance}} = \frac{k}{2N - k}
\end{equation}

and report \emph{lift} as the ratio of observed to chance Jaccard:
\begin{equation}
    \text{Lift}_k = \frac{J_k^{\text{observed}}}{J_k^{\text{chance}}}
\end{equation}

Lift $= 1$ indicates rankings no better than random; lift $> 1$ indicates that the same features tend to appear in the top-$k$ set 
across conditions more than expected by chance. We report mean $J_k$ and lift across all condition pairs for $k \in \{10, 20, 30, 
40, 50\}$.

\subsection{Cohen's d effect sizes}
Unlike within-architecture experiments where the same features are observed across conditions, cross-architecture comparison does not permit per-feature reliability analysis: features are not matched across architectures and ICC or Jaccard metrics are therefore not applicable. We instead compare score \textit{distributions} via mean scores and Cohen's $d$ effect sizes. Formally, for metric $m$ and architectures $A$ and $B$, we compute:

\begin{equation}
    d = \frac{\bar{x}_{m,A} - \bar{x}_{m,B}}{s_{\text{pooled}}}, \quad s_{\text{pooled}} = \sqrt{\frac{(n_A - 1)s_{m,A}^2 + (n_B - 1)s_{m,B}^2}{n_A + n_B - 2}}
\end{equation}

where $\bar{x}_{m,A}$, $s_{m,A}$, and $n_A$ denote the sample mean, standard deviation, and number of features for architecture $A$, respectively. The small and inconsistent effect sizes observed do not prove that no architectural differences exist at the feature level; they show only that differences are not detectable at the distribution level with our sample sizes. This is nonetheless informative, because if architectural differences were large, they would be expected to show as a distributional shift that is detectable even without per-feature correspondence.

\subsection{Variance Decomposition}
\label{app:variance_decomp}

To quantify the contribution of each pipeline choice to metric variability, we fit a variance components model via Restricted Maximum Likelihood (REML). For score $s_{ijklm}$ of feature $i$ under corpus $j$, draw $k$, explainer $l$, and architecture $m$:

\begin{equation}
    s_{ijklm} = \mu + \alpha_i + \beta_j + \gamma_k + \delta_l + \epsilon_m 
    \label{eq:variance_model}
\end{equation}

where $\mu$ is the grand mean and $\alpha_i, \beta_j, \gamma_k, \delta_l, \epsilon_m$ are independent mean-zero random effects for feature identity, corpus, draw, explainer, and architecture respectively:

\begin{equation}
    \alpha_i \sim \mathcal{N}(0, \sigma^2_\alpha), \quad
    \beta_j \sim \mathcal{N}(0, \sigma^2_\beta), \quad
    \gamma_k \sim \mathcal{N}(0, \sigma^2_\gamma), \quad
    \delta_l \sim \mathcal{N}(0, \sigma^2_\delta), \quad
    \epsilon_m \sim \mathcal{N}(0, \sigma^2_\epsilon)
\end{equation}

The total variance decomposes as:

\begin{equation}
    \sigma^2_{\text{total}} = \sigma^2_\alpha + \sigma^2_\beta + \sigma^2_\gamma + \sigma^2_\delta + \sigma^2_\epsilon 
\end{equation}

The proportion of variance attributable to each factor $f$ is then:

\begin{equation}
    \rho_f = \frac{\sigma^2_f}{\sigma^2_{\text{total}}}
\end{equation}

Variance components are estimated per metric from partially overlapping subsets, as not all pipeline combinations are fully crossed.

\section{SAE Training Details}
\label{app:sae_training_details}
\subsection{Pythia SAE models}
\label{app:pythiadetails}
There are three pretrained SAEs accessed in this study \cite{karvonen2025saebench}. For most experiments, we opt for BatchTopK SAE with \textit{k} = 160. For the experiment comparing architectures, we additionally investigate a Matryoshka SAE with \textit{k} = 160 and a ReLU SAE with a sparsity penalty of 0.03. All SAEs can be found at \verb|adamkarvonen/saebench_pythia-160m-deduped_width-2pow14_date-0108|. Select the folders \verb|trainer_3|. The three pretrained SAEs from SAEBench \citep{karvonen2025saebench} that are 
used in this study are all trained on layer 8 of \textsc{Pythia-160M}. Table~\ref{tab:sae_hyperparams} summarises their hyperparameters.

\begin{table}[h]
\centering
\caption{SAE hyperparameters for Pythia-160M models from SAEBench. 
All models available at \texttt{adamkarvonen/saebench\_pythia-160m-deduped\_width-2pow14\_date-0108}.}
\label{tab:sae_hyperparams}
\small
\begin{tabular}{llccc}
\toprule
\textbf{Architecture} & \textbf{Trainer} & \textbf{$k$ / penalty} 
& \textbf{Dict.\ size} & \textbf{Layer} \\
\midrule
BatchTopK   & \texttt{trainer\_3} & $k = 160$  & $2^{14} = 16{,}384$ &  8\\
Matryoshka  & \texttt{trainer\_3}    & $k = 160$ 
  & $2^{14} = 16{,}384$ & 8  \\
Standard    & \texttt{trainer\_5}    & $\lambda = 0.03$ 
  & $2^{14} = 16{,}384$ & 8  \\
\bottomrule
\end{tabular}
\end{table}

For most experiments we use the BatchTopK SAE with $k = 160$ as the 
primary evaluation target. For the architecture comparison 
(Exp.\ 3), we additionally evaluate the Matryoshka SAE and Standard 
(ReLU) SAE. 

The time  for one experiment on \textsc{Pythia-160M} is roughly . The experiments can be executed on a CPU. 

\subsection{Apertus SAE Training Details}
\label{app:apertus_sae}

This paper is the first to train and evaluate SAEs on Apertus 
\citep{swissai2025apertus}, a Swiss multilingual large language model trained on German, French, Italian, Romansh, and English text. SAEs were trained on layer 16 of Apertus-8B using activations collected from the Pile (uncopyrighted subset). \autoref{tab:apertus_sae_hyperparams} 
summarises the training hyperparameters.

The size of Matryoshka is slightly smaller compared to the ReLU and BatchTopK variant. This was done with efficiency and computational limitatios

\begin{table}[h]
\centering
\caption{Apertus-8B SAE training hyperparameters across the three architectures evaluated. All SAEs are trained on activations from layer 16 of Apertus-8B (residual stream, $d=4{,}096$) using the Pile (uncopyrighted) corpus.}
\label{tab:apertus_sae_hyperparams}
\small
\begin{tabular}{lccc}
\toprule
\textbf{Hyperparameter} & \textbf{BatchTopK} & \textbf{Matryoshka} & \textbf{Standard} \\
\midrule
Architecture        & BatchTopK & Matryoshka (BatchTopK) & ReLU + L1 \\
Layer               & 16 & 16 & 16 \\
Residual stream dim & 4{,}096 & 4{,}096 & 4{,}096 \\
Dictionary size     & 65{,}536 ($\times$16) & 51{,}200 ($\times$12.5) & 65{,}536 ($\times$16) \\
Matryoshka groups   & --- & [6{,}144, 14{,}336, 30{,}720] & --- \\
Sparsity mechanism  & BatchTopK ($k=160$) & BatchTopK ($k=160$) & L1 ($\lambda = 4 \times 10^{-4}$) \\
Training steps      & 140{,}000 & 300{,}000 & 200{,}000 \\
Batch size          & 4{,}096 & 4{,}096 & 4{,}096 \\
Learning rate       & 0.001 & 0.001 & 0.001 \\
LR warmup steps     & 1{,}000 & 1{,}000 & 1{,}000 \\
LR decay            & $\times$0.8 every 10{,}000 & $\times$0.9 every 20{,}000 & $\times$0.8 every 10{,}000 \\
Minimum LR          & $1 \times 10^{-6}$ & $1 \times 10^{-6}$ & $1 \times 10^{-6}$ \\
Auxiliary penalty   & 0.03125 & 0.03125 & --- \\
Top-$k$ auxiliary   & 128 & 128 & --- \\
Dead latent window  & 50 batches & 50 batches & 50 batches \\
\bottomrule
\end{tabular}
\end{table}

A key difference between Apertus and Pythia-160M is the activation scale: Apertus features activate with magnitudes in the hundreds to thousands, compared to single digits for Pythia-160M. The activation threshold for feature inclusion is therefore set to 100 for Apertus (vs.\ 0.1 for Pythia-160M). Figure~\ref{fig:activation_diagnostic} confirms that both thresholds yield comparable activation coverage within each model.

The large dictionary size ($d_{\text{SAE}} = 65{,}536$) combined with batch-level sparsity enforcement ($k = 160$) results in an 
expected per-feature activation rate of approximately:
\begin{equation}
    P(\text{feature activates}) = \frac{k}{d_{\text{SAE}}} 
    = \frac{160}{65{,}536} \approx 0.24\%
\end{equation}
per token. Over a typical evaluation sequence of 32 tokens, a randomly sampled feature is therefore expected to activate on fewer than one token per sequence on average, which explains the low feature evaluability observed for BatchTopK on Apertus (Table~\ref{tab:feature_coverage}). Therefore, we randomly sample 2000 features to extract around 250 activating features. 

\textcolor{insightone}{Due to anonymisation, we removed the link of the trained SAEs, but the models are already released.}

\subsubsection{Training and Evaluation Scores}
As these SAEs were trained by us, we report the train and evaluation scores. The best checkpoint was taken as the final model checkpoint.

\begin{table}[ht]
\centering
\caption{SAE training and evaluation metrics across configurations.}
\label{tab:sae_metrics}
\begin{tabular}{lcccccc}
\toprule
& \multicolumn{2}{c}{\textbf{Rel.\ RMSE} $\downarrow$} & \multicolumn{2}{c}{\textbf{Loss Recovered} $\uparrow$} & \multicolumn{2}{c}{\textbf{Feature Sparsity} $\uparrow$} \\
\cmidrule(lr){2-3} \cmidrule(lr){4-5} \cmidrule(lr){6-7}
\textbf{SAE} & Train & Eval & Train & Eval & Train & Eval \\
\midrule
ReLU (L16)         & 0.00111 & 0.00107 & 0.99999 & 0.99999 & 0.9377 & 0.9377 \\
BatchTopK $k$=160  & 0.04125 & 0.04124 & 0.99830 & 0.99829 & 0.9976 & 0.9976 \\
Matryoshka $k$=160 & 0.03345 & 0.04176 & 0.99888 & 0.99824 & 0.9969 & 0.9968 \\
\bottomrule
\end{tabular}
\end{table}

\subsection{Time and Compute}
\label{app:compute}
The differences in LLM calls for the two models stems from the fact that not every condition has the exact amount of features, even though we try in both cases to reach roughly 250 features. 

\begin{table}[H]
\caption{\textsc{Pythia-160M}: LLM calls, time and compute estimates. The code is executing on a CPU. }
  \centering
  \begin{tabular}{lrr}
    \toprule
    \textbf{Experiment} & \textbf{LLM Calls} & \textbf{Time (h)} \\
    \midrule
    Exp1: Cross-Dataset Generalization & 4,705 & 4.7 \\
    Exp2a: Sampling Instability & 5,970 & 6.0 \\
    Exp2b ($w{=}5$): Cross-Explainer & 2,490 & 2.5 \\
    Exp2b ($w{=}10$): Cross-Explainer & 2,670 & 2.7 \\
    Exp2c: Paraphrase Stability & 4,165 & 4.2 \\
    Exp4: Cross-SAE Architecture & 3,665 & 3.7 \\
    \midrule
    \textbf{Total} & \textbf{23,665} & \textbf{23.7} \\
    \bottomrule
  \end{tabular}
  \label{tab:compute_pythia}
\end{table}

\begin{table}[H]
\caption{\textsc{Apertus-8B}: LLM calls, time and compute estimates. The code is executed using GH200 GPU on a supercomputer. The latency originates from the LLM calls, not from the GPU. }
  \centering
  \begin{tabular}{lrr}
    \toprule
    \textbf{Experiment} & \textbf{LLM Calls} & \textbf{Time (h)} \\
    \midrule
    Exp1: Cross-Dataset Generalization & 5,035 & 5.1 \\
    Exp2a: Sampling Instability & 3,975 & 4.0 \\
    Exp2b ($w{=}5$): Cross-Explainer & 1,780 & 1.8 \\
    Exp2b ($w{=}10$): Cross-Explainer & 2,550 & 2.6 \\
    Exp2c: Paraphrase Stability & 5,120 & 5.1 \\
    Exp4: Cross-SAE Architecture & 4,190 & 4.2 \\
    \midrule
    \textbf{Total} & \textbf{22,650} & \textbf{27.4} \\
    \bottomrule
  \end{tabular}
  \label{tab:compute_apertus}
\end{table}

\section{Extended Results}

\subsection{Similarity Values}
\label{app:similarity_values}
\textit{Cosine similarity and \textsc{BERTScore} of explanations remain high and narrowly distributed across all conditions, even when scores vary 
substantially, invalidating explanation similarity as a proxy for stability.} 

Trivially, to combat instability researchers could monitor explanation similarity as a cheap diagnostic: if two explanations are semantically close, their scores should agree. \autoref{tab:similarity} falsifies this assumption systematically across all four experiments and both models. Mean cosine similarity ranges from $0.840$ and $0.824$ for (\textsc{Corpus}, $0.942$ to $0.949$ for (\textsc{Paraphrase}, and the ordering \textsc{Paraphrase} $>$ \textsc{Draw} $>$ \textsc{Explainer} $\approx$ \textsc{Corpus} holds consistently. However, the score instability does not follow this ordering ({tab:icc}). \textsc{Draw} produces the most severe instability (negative ICC for \textsc{Simulation} and \textsc{Purity}) while maintaining high explanation similarity ($0.892$ cosine on \texttt{Pythia-160M}).

The dissociation is most striking for \textsc{Draw} on feature 2208 (\autoref{app:qual_examples}), where all five draws produce near-identical explanations (cosine $= 0.964$, BERTScore $= 0.944$), yet \textsc{Simulation} scores range from $-0.177$ to $+0.729$. \textsc{Detection} remains fixed at $0.650$ throughout. The explanation is essentially the same; while the score differs. This rules out explanation drift as the mechanism behind instability under draw variation, and points instead to the held-out evaluation sequences themselves as the source of noise.

Explanation similarity therefore cannot substitute for the direct stability measurement provided by ICC and Jaccard lift. A pipeline that produces consistent explanations may still produce unreliable scores, and this failure is invisible without running the stability checks we propose in \autoref{sec:recommendations}.

\begin{table}[h!]
\centering
\caption{Mean pairwise explanation similarity per experiment and model. Cosine 
similarity computed using \texttt{E5-small-v2}; BERTScore using 
\texttt{RoBERTa-large}. Both reported as mean $\pm$ std across all pairwise 
comparisons within each experiment.}
\label{tab:similarity}
\footnotesize 
\begin{tabular}{llrcccc}
\toprule
& & & \multicolumn{2}{c}{\textbf{Cosine similarity}} 
    & \multicolumn{2}{c}{\textbf{BERTScore O1}} \\
\cmidrule(lr){4-5} \cmidrule(lr){6-7}
\textbf{Model} & \textbf{Experiment} & \textbf{$n$} 
  & \textbf{Mean} & \textbf{Std} 
  & \textbf{Mean} & \textbf{Std} \\
\midrule
\multirow{4}{*}{\texttt{Pythia-160M}}
  & \textsc{Corpus}     & 258 & 0.840 & 0.028 & 0.844 & 0.020 \\
  & \textsc{Draw}       & 239 & 0.892 & 0.036 & 0.874 & 0.031 \\
  & \textsc{Explainer}  & 266 & 0.853 & 0.036 & 0.849 & 0.022 \\
  & \textsc{Paraphrase} & 267 & 0.942 & 0.037 & 0.924 & 0.038 \\
\midrule
\multirow{4}{*}{\texttt{Apertus-8B}}
  & \textsc{Corpus}     & 264 & 0.824 & 0.024 & 0.832 & 0.019 \\
  & \textsc{Draw}       & 159 & 0.878 & 0.029 & 0.864 & 0.026 \\
  & \textsc{Explainer}  & 256 & 0.841 & 0.034 & 0.841 & 0.020 \\
  & \textsc{Paraphrase} & 256 & 0.949 & 0.022 & 0.924 & 0.016 \\
\bottomrule
\end{tabular}
\end{table}

\subsection{Variance Decomposition values}
\label{app:variance_decomp_table}

\begin{table}[H]
\centering
\caption{Variance decomposition (\% of total variance) per source and metric, 
for \texttt{Pythia-160M} and \texttt{Apertus-8B}, with 95\% confidence 
intervals in brackets. The dominant source per (model, metric) is shown in 
\textbf{bold}.}
\label{tab:variance_decomp}
\small
\resizebox{\textwidth}{!}{%
\begin{tabular}{llccccc}
\toprule
\textbf{Model} & \textbf{Metric}
  & $\boldsymbol{\sigma^2_{\mathrm{corpus}}}$
  & $\boldsymbol{\sigma^2_{\mathrm{draw}}}$
  & $\boldsymbol{\sigma^2_{\mathrm{explainer}}}$
  & $\boldsymbol{\sigma^2_{\mathrm{paraphrase}}}$
  & $\boldsymbol{\sigma^2_{\mathrm{arch}}}$ \\
\midrule
\multirow{4}{*}{\texttt{Pythia-160M}}
  & \textsc{Simulation} & 38 [29, 48] & \textbf{50 [46, 54]} & 5 [4, 7]    & 4 [2, 6]    & 2 [1, 5]  \\
  & \textsc{Detection}  & \textbf{60 [47, 72]} & 20 [16, 24] & 11 [6, 18]  & 8 [5, 12]   & 1 [0, 3]  \\
  & \textsc{Fuzzing}    & \textbf{48 [32, 68]} & 11 [10, 13] & 11 [8, 15]  & 29 [15, 46] & 0 [0, 2]  \\
  & \textsc{Purity}     & 24 [20, 28] & \textbf{63 [57, 68]} & 7 [5, 9]    & 7 [5, 10]   & 0 [0, 1]  \\
\midrule
\multirow{4}{*}{\texttt{Apertus-8B}}
  & \textsc{Simulation} & 39 [31, 48] & \textbf{47 [43, 52]} & 5 [4, 7]    & 5 [3, 8]    & 3 [1, 5]  \\
  & \textsc{Detection}  & \textbf{46 [37, 57]} & 17 [15, 19] & 9 [7, 10]   & 13 [8, 20]  & 15 [9, 23] \\
  & \textsc{Fuzzing}    & 29 [17, 44] & 8 [8, 9]    & 6 [5, 7]    & \textbf{57 [38, 80]} & 0 [0, 1]  \\
  & \textsc{Purity}     & 23 [20, 26] & \textbf{62 [56, 68]} & 9 [7, 11]   & 5 [4, 7]    & 1 [1, 3]  \\
\bottomrule
\end{tabular}
}
\end{table}

\subsection{ICC Heatmap}
\label{sec:icc_heatmap}
\begin{figure}[H]
\includegraphics[width=1.0\textwidth]{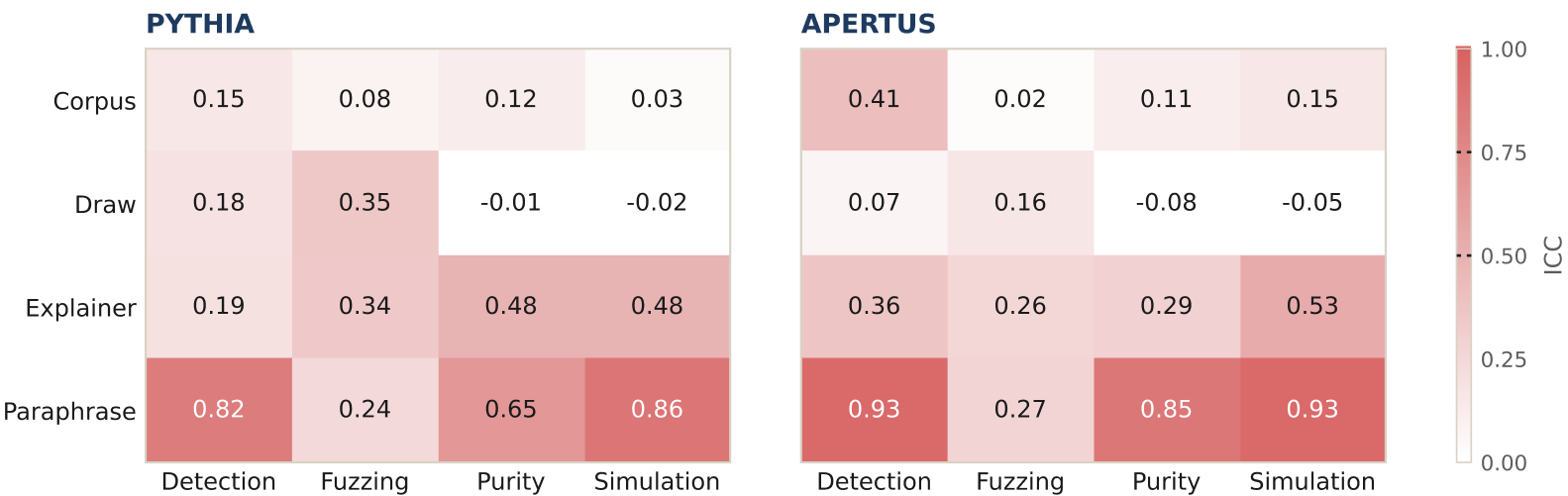}
\caption{ICC heatmap with the delta of the mean for each corpus and metric }
\centering
 \label{fig:icc_heatmap}
\end{figure}

\subsection{Mean scores and standard deviation per condition}

\begin{table}[H]
\centering
\caption{Mean scores and standard deviations per condition, metric, and model across all experiments. As mentioned earlier, the mean and variance cover up the instability of different conditions. }
\label{tab:mean_scores}
\scriptsize
\setlength{\tabcolsep}{4pt}
\begin{tabular}{llrrrrrrrr}
\toprule
& & \multicolumn{2}{c}{\textbf{Simulation}} 
  & \multicolumn{2}{c}{\textbf{Detection}} 
  & \multicolumn{2}{c}{\textbf{Fuzzing}} 
  & \multicolumn{2}{c}{\textbf{Purity}} \\
\cmidrule(lr){3-4}\cmidrule(lr){5-6}\cmidrule(lr){7-8}\cmidrule(lr){9-10}
\textbf{Exp.} & \textbf{Condition} 
  & \textbf{mean} & \textbf{std} 
  & \textbf{mean} & \textbf{std} 
  & \textbf{mean} & \textbf{std} 
  & \textbf{mean} & \textbf{std} \\
\midrule
\multicolumn{10}{l}{\textit{Pythia-160M}} \\
\midrule
\multirow{4}{*}{Corpus}
  & Pile      & 0.705 & 0.255 & 0.655 & 0.086 & 0.559 & 0.091 & 0.628 & 0.110 \\
  & C4        & 0.674 & 0.302 & 0.673 & 0.113 & 0.572 & 0.091 & 0.628 & 0.115 \\
  & Wikipedia & 0.663 & 0.310 & 0.671 & 0.116 & 0.567 & 0.098 & 0.623 & 0.123 \\
  & Code      & 0.567 & 0.415 & 0.688 & 0.136 & 0.533 & 0.097 & 0.621 & 0.146 \\
\midrule
\multirow{5}{*}{Draw}
  & Draw 0    & 0.206 & 0.347 & 0.611 & 0.053 & 0.536 & 0.043 & 0.554 & 0.208 \\
  & Draw 1    & 0.273 & 0.292 & 0.617 & 0.057 & 0.538 & 0.044 & 0.623 & 0.199 \\
  & Draw 2    & 0.522 & 0.244 & 0.629 & 0.054 & 0.543 & 0.046 & 0.485 & 0.102 \\
  & Draw 3    & 0.081 & 0.212 & 0.618 & 0.063 & 0.534 & 0.045 & 0.394 & 0.109 \\
  & Draw 4    & $-$0.124 & 0.228 & 0.616 & 0.051 & 0.542 & 0.054 & 0.302 & 0.122 \\
\midrule
\multirow{2}{*}{Explainer}
  & \texttt{Gemini Flash}    & 0.740 & 0.187 & 0.643 & 0.058 & 0.555 & 0.055 & 0.621 & 0.106 \\
  & \texttt{GPT4o-mini}& 0.753 & 0.174 & 0.637 & 0.047 & 0.550 & 0.051 & 0.615 & 0.098 \\
\midrule
\multirow{4}{*}{Paraphrase}
  & Original    & 0.670 & 0.295 & 0.671 & 0.103 & 0.563 & 0.108 & 0.637 & 0.107 \\
  & Paraphrase 1 & 0.649 & 0.314 & 0.663 & 0.114 & 0.565 & 0.090 & 0.610 & 0.139 \\
  & Paraphrase 2 & 0.635 & 0.323 & 0.660 & 0.111 & 0.567 & 0.089 & 0.605 & 0.157\\
  & Paraphrase 3 & 0.634 & 0.325 & 0.663 & 0.110 & 0.563 & 0.091 & 0.606 & 0.154 \\
\midrule
\multirow{3}{*}{Architecture}
  & BatchTopK   & 0.682 & 0.302 & 0.657 & 0.095 & 0.554 & 0.104 & 0.613 & 0.130 \\
  & Matryoshka  & 0.602 & 0.435 & 0.678 & 0.126 & 0.546 & 0.110 & 0.622 & 0.131 \\
  & Standard    & 0.756 & 0.221 & 0.658 & 0.083 & 0.545 & 0.066 & 0.633 & 0.095 \\
\midrule
\multicolumn{10}{l}{\textit{Apertus-8B}} \\
\midrule
\multirow{2}{*}{Corpus}
  & Pile      & 0.532 & 0.418 & 0.694 & 0.146 & 0.523 & 0.115 & 0.618 & 0.153 \\
  & C4        & 0.517 & 0.426 & 0.680 & 0.139 & 0.515 & 0.091 & 0.583 & 0.148 \\
\midrule
\multirow{5}{*}{Draw}
  & Draw 0    & 0.151 & 0.329 & 0.604 & 0.051 & 0.520 & 0.036 & 0.559 & 0.216 \\
  & Draw 1    & 0.205 & 0.276 & 0.599 & 0.049 & 0.523 & 0.039 & 0.610 & 0.223 \\
  & Draw 2    & 0.459 & 0.306 & 0.600 & 0.047 & 0.522 & 0.036 & 0.460 & 0.114 \\
  & Draw 3    & 0.085 & 0.226 & 0.603 & 0.047 & 0.526 & 0.037 & 0.398 & 0.117 \\
  & Draw 4    & $-$0.128 & 0.262 & 0.601 & 0.050 & 0.524 & 0.037 & 0.283 & 0.093 \\
\midrule
\multirow{2}{*}{Explainer}
  & \texttt{Gemini Flash}      & 0.672 & 0.225 & 0.617 & 0.052 & 0.520 & 0.040 & 0.564 & 0.122\\
  & \texttt{GPT4o-mini}& 0.709 & 0.201 & 0.623 & 0.053 & 0.525 & 0.044 & 0.589 & 0.108 \\
\midrule
\multirow{4}{*}{Paraphrase}
  & Original    & 0.492 & 0.437 & 0.703 & 0.157 & 0.526 & 0.107 & 0.629 & 0.157 \\
  & Paraphrase 1 & 0.482 & 0.463 & 0.700 & 0.158 & 0.522 & 0.101 & 0.619 & 0.159 \\
  & Paraphrase 2 & 0.494 & 0.456 & 0.698 & 0.159 & 0.528 & 0.107 & 0.624 & 0.160 \\
  & Paraphrase 3 & 0.501 & 0.449 & 0.699 & 0.156 & 0.514 & 0.106 & 0.624 & 0.163 \\
\midrule
\multirow{3}{*}{Architecture}
  & BatchTopK   & 0.508 & 0.436 & 0.700 & 0.152 & 0.523 & 0.123 & 0.623 & 0.151 \\
  & Matryoshka  & 0.675 & 0.241 & 0.619 & 0.055 & 0.525 & 0.040 & 0.575 & 0.123 \\
  & Standard    & 0.637 & 0.259 & 0.618 & 0.068 & 0.527 & 0.039 & 0.554 & 0.138 \\
\bottomrule
\end{tabular}
\end{table}

\subsection{Corpus Results}

\begin{figure}[H]
\includegraphics[width=1.0\textwidth]{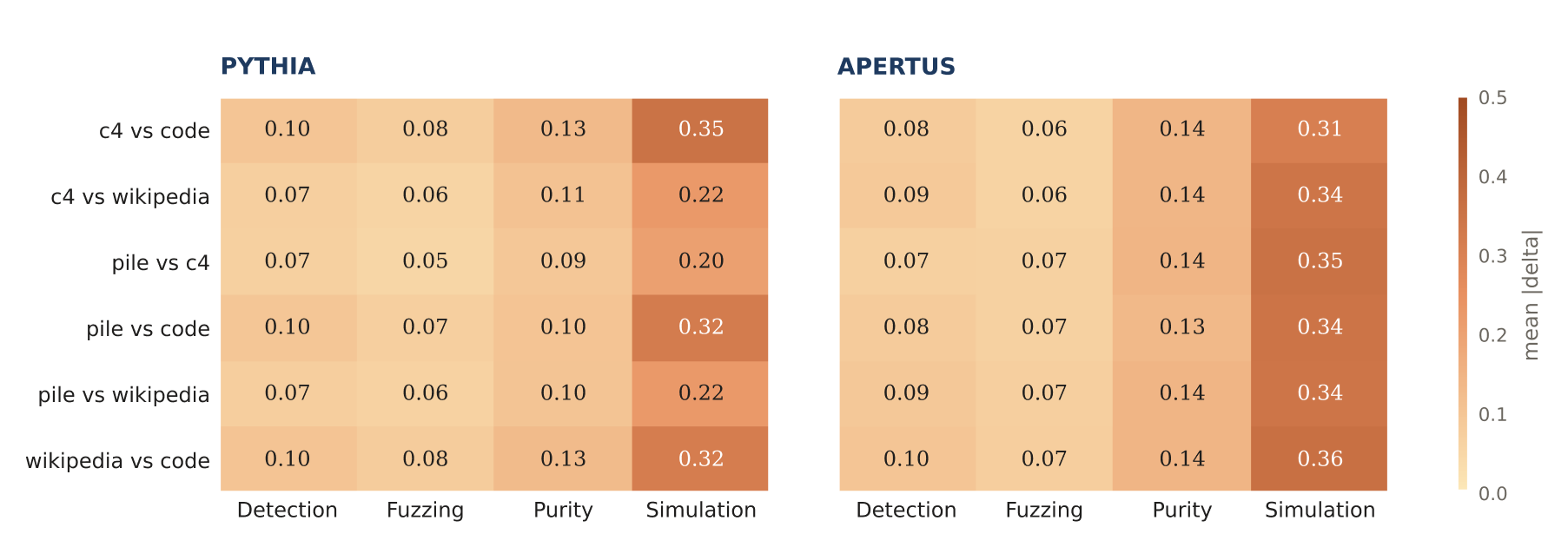}
\caption{Corpus heatmap with the delta of the mean for each corpus combination and metric }
\centering
 \label{fig:delta_heatmap}
\end{figure}

\autoref{fig:delta_heatmap} shows the delta in scores between each condition. This shows that most of the deltas are relatively similar, even when the datasets differ. The expectation was that more divergent datasets would yield higher deltas, but this does not appear to be the case. This could be due to the inherent design of the metric or due to the possible polysemanticity of features. 

\label{sec:corpus_results}


\subsection{Jaccard Lift Table}

\begin{table}[H]
\centering
\caption{Top-$k$ Jaccard lift (observed/chance) per metric and source of 
variation at $k=10$ and $k=20$, for \texttt{Pythia-160M} and 
\texttt{Apertus-8B}, with 95\% confidence intervals in brackets. 
Lift $= 1\times$ indicates feature selection no better than random. 
Values below $2\times$ are shown in \textbf{bold} to indicate near-chance 
selection. \footnotesize{$^\dagger$ Point estimate falls outside BCa (bias-corrected and accelerated) CI due to 
skewness in the pairwise Jaccard distribution; the CI should be 
interpreted as the primary reliability indicator.}}
\label{tab:jaccard_lift}
\renewcommand{\arraystretch}{1.8}
\resizebox{\textwidth}{!}{%
\begin{tabular}{llcccc}
\toprule
\multicolumn{6}{c}{\normalsize\textbf{\texttt{Pythia-160M}}} \\
\midrule
\textbf{Source} & $k$ 
  & \textbf{\textsc{Simulation}} & \textbf{\textsc{Detection}} 
  & \textbf{\textsc{Fuzzing}} & \textbf{\textsc{Purity}} \\
\midrule
\multirow{2}{*}{\textsc{Corpus}}
  & 10 & \textbf{0.68$\times$ [0.00, 2.63]} & 4.34$\times$ [2.85, 10.01] & 2.81$\times$ [1.00, 4.03] & 3.85$\times$ [2.09, 8.53] \\
  & 20 & \textbf{1.21$\times$ [0.80, 2.05]} & \textbf{1.60$\times$ [0.96, 2.55]} & 2.61$\times$ [1.42, 3.32] & 3.52$\times$ [2.08, 5.08] \\
\midrule
\multirow{2}{*}{\textsc{Draw}}
  & 10 & \textbf{0.82$\times$ [0.00, 1.78]} & 2.09$\times$ [0.00, 3.21] & 6.19$\times$ [3.46, 5.72]$^\dagger$ & \textbf{1.68$\times$ [0.41, 5.33]} \\
  & 20 & \textbf{1.43$\times$ [0.59, 2.37]} & \textbf{1.76$\times$ [0.90, 2.82]} & 3.80$\times$ [2.68, 3.62]$^\dagger$ & \textbf{1.35$\times$ [0.54, 1.94]} \\
\midrule
\multirow{2}{*}{\textsc{Explainer}}
  & 10 & 12.15$\times$ [0.88, 20.83] & 2.56$\times$ [0.00, 8.58] & 8.58$\times$ [0.00, 16.20] & 20.83$\times$ [8.58, 39.76] \\
  & 20 & 9.03$\times$ [5.05, 14.28] & 4.20$\times$ [2.64, 7.87] & 9.03$\times$ [5.95, 15.62] & 11.46$\times$ [6.91, 16.52] \\
\midrule
\multirow{2}{*}{\textsc{Paraphrase}}
  & 10 & 28.22$\times$ [9.25, 42.87] & 42.87$\times$ [34.93, 52.40] & 9.25$\times$ [0.00, 22.46] & 42.87$\times$ [22.46, 52.40] \\
  & 20 & 12.37$\times$ [6.28, 17.13] & 19.00$\times$ [12.37, 23.25] & 9.75$\times$ [6.42, 15.42] & 21.03$\times$ [15.42, 25.70] \\
\midrule
\multicolumn{6}{c}{\large\textbf{\texttt{Apertus-8B}}} \\
\midrule
\textbf{Source} & $k$ 
  & \textbf{\textsc{Simulation}} & \textbf{\textsc{Detection}} 
  & \textbf{\textsc{Fuzzing}} & \textbf{\textsc{Purity}} \\
\midrule
\multirow{2}{*}{\textsc{Corpus}}
  & 10 & \textbf{0.97$\times$ [0.00, 2.01]} & 9.22$\times$ [5.09, 14.42] & \textbf{1.65$\times$ [0.32, 2.50]} & 4.57$\times$ [1.97, 8.22] \\
  & 20 & \textbf{1.23$\times$ [0.64, 2.32]} & 5.35$\times$ [3.52, 7.85] & \textbf{1.73$\times$ [1.08, 2.61]} & 2.22$\times$ [1.13, 2.61] \\
\midrule
\multirow{2}{*}{\textsc{Draw}}
  & 10 & \textbf{0.97$\times$ [0.32, 1.62]} & \textbf{0.99$\times$ [0.18, 1.54]} & 4.52$\times$ [2.90, 8.63] & 2.22$\times$ [1.03, 3.70] \\
  & 20 & \textbf{0.85$\times$ [0.55, 1.08]} & \textbf{1.81$\times$ [1.13, 2.44]} & 2.68$\times$ [1.79, 3.55] & \textbf{1.29$\times$ [0.81, 1.85]} \\
\midrule
\multirow{2}{*}{\textsc{Explainer}}
  & 10 & 6.07$\times$ [1.81, 14.74] & 6.07$\times$ [1.81, 18.52] & 8.60$\times$ [6.07, 14.74] & 3.82$\times$ [0.00, 6.07] \\
  & 20 & 3.54$\times$ [1.86, 7.16] & 3.54$\times$ [1.35, 6.33] & 2.95$\times$ [1.35, 5.57] & 3.54$\times$ [0.98, 5.57] \\
\midrule
\multirow{2}{*}{\textsc{Paraphrase}}
  & 10 & 43.76$\times$ [33.39, 49.80] & 43.76$\times$ [35.29, 49.80] & 12.07$\times$ [7.59, 22.78] & 30.20$\times$ [15.15, 38.23] \\
  & 20 & 15.47$\times$ [11.16, 19.73] & 21.72$\times$ [19.67, 24.40] & 7.17$\times$ [5.61, 10.50] & 19.03$\times$ [15.34, 22.46] \\
\bottomrule
\end{tabular}
}
\end{table}

\subsection{Draw Score Instability}
\label{app:draw_instability}
\begin{figure}[H]
\centering
\includegraphics[width=\textwidth]{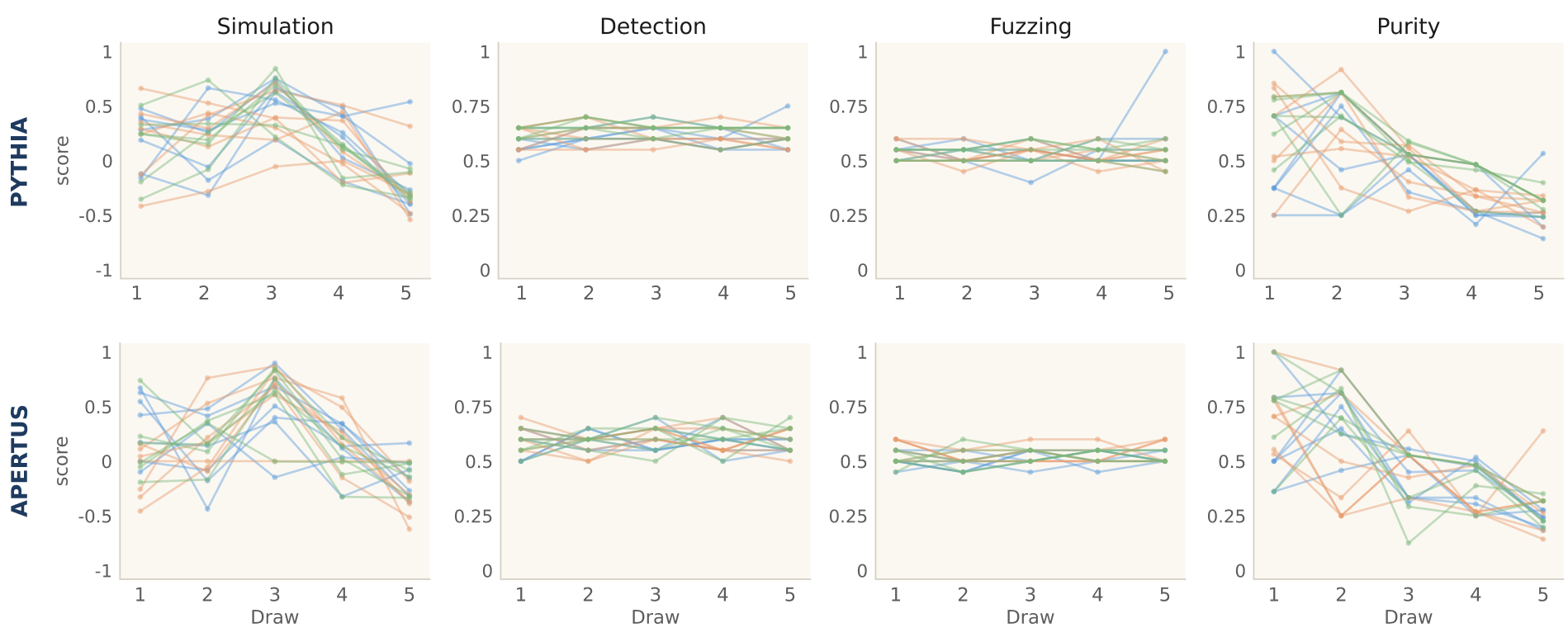}
\caption{The score across draws for each feature. One line represents one feature. Both \textsc{Simulation} and \textsc{Purity} show a high variation in scores for each draw.}
\label{fig:draw_instability}
\end{figure}

\textsc{Draw} contributes to a large portion of the variance, as we saw in \autoref{sec:findings_overview}. Below, we provide a figure that clearly shows the trajectory of 20 randomly picked features for the scores per draw for each feature (\autoref{fig:draw_instability}).

\subsection{Explainer Sensitivity}
\label{app:explainer_sensitivity}
To test whether \texttt{GPT4o-mini}'s tendency to produce longer explanations confounds the comparison, we additionally ran a length-constrained condition in which both explainers were restricted to a maximum of ten words. 

\begin{figure}[H]
\centering
\includegraphics[width=\textwidth]{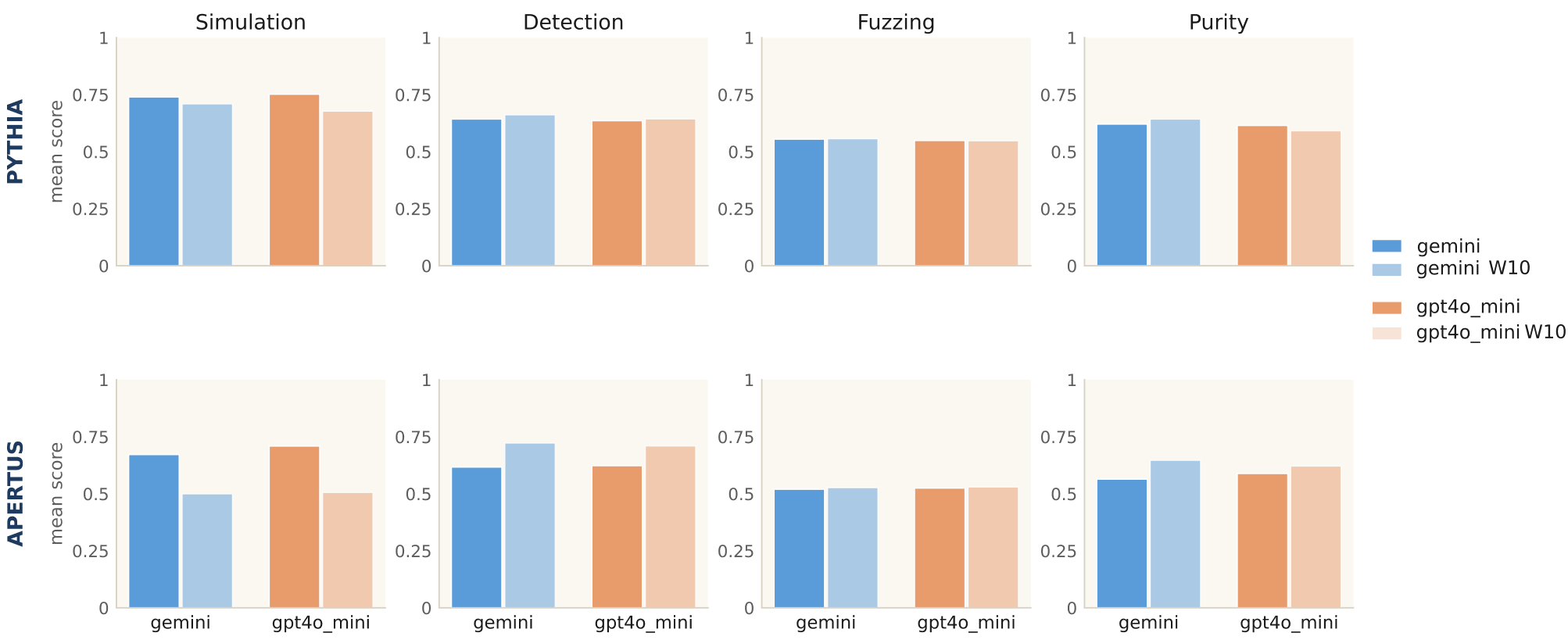}
\caption{Mean metric scores for unrestricted and W10 
(10-word-constrained) explanations, per explainer and model. 
Restricting explanation length reduces Simulation scores for both explainers but has minimal effect on Detection and Fuzzing.}
\label{fig:compare_orig_vs_w10}
\end{figure}

Figure~\ref{fig:compare_orig_vs_w10} compares mean scores between unrestricted and length-constrained conditions for both explainers. Restricting explanation length to ten words reduces \textsc{Simulation} and \textsc{Detection} scores for both explainers on \texttt{Apertus-8B} but has minimal effect on \texttt{Pythia-160M}. \textsc{Fuzzing} shows a small differences between explainers for both models, which could indicate that the score is least sensitive to explanation detail, compared to other metrics.


\subsection{Paraphrase Sensitivity}
\label{app:paraphrase_sensitivity}

Mean pairwise cosine similarity across the four versions (original 
$+$ three paraphrases) is 0.942 for Pythia-160M and 0.949 for Apertus-8B (Table~\ref{tab:similarity}), which is substantially higher than cross-explainer similarity ($\approx$0.853) or cross-corpus similarity ($\approx$0.841).

\begin{figure}[h!]
\centering
\includegraphics[width=\textwidth]{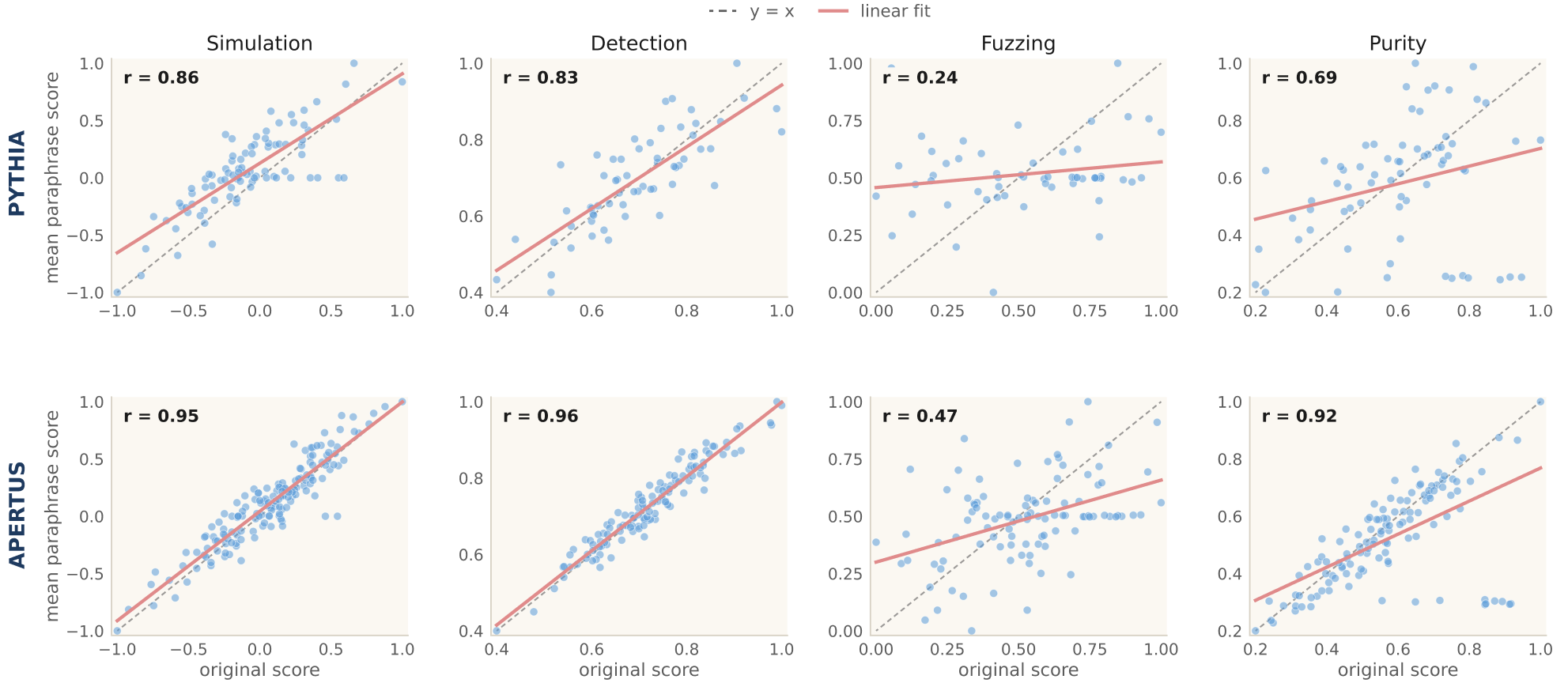}
\caption{The score of the original paraphrase vs. the average of the other paraphrases. }
\label{fig:paraphrase_average}
\end{figure}

\autoref{fig:trajectory_explainer_scatter} shows per-feature score trajectories across the original explanation and three paraphrases. The trajectories of all metrics show substantial crossing and non-monotonic patterns despite the high explanation similarity (mean cosine $= 0.945$, \autoref{tab:similarity}). Next to that, \autoref{fig:paraphrase_average} shows the score of the original paraphrase vs. the mean score of the alternative paraphrases. The lines for \textsc{simulation} and \textsc{detection} are relatively similar to $y=x$, but varies the most for \textsc{fuzzing}.

\subsubsection{Paraphrase Generator Model}
For the original experiment, we keep the explainer, scorer and paraphrases model fixed to \texttt{Gemini Flash}. However, we conducted a small experiment, by changing the paraphrase generator to \texttt{GPT4o-mini}, to observe the effects of the paraphase model. 

\autoref{tab:paraphrase_exp} reports ICC(2,1) and Jaccard overlap at top-10 and top-20 thresholds for two paraphrase models across four metrics. Overall, \textsc{simulation} and \textsc{detection} scores show high reliability under paraphrasing, with ICC(2,1) values exceeding 0.8 for both \texttt{GPT4o-mini} and \texttt{Gemini Flash}. \textsc{Fuzzing} exhibits substantially lower agreement, with an ICC(2,1) of 0.328 for \texttt{GPT4o-mini} and a confidence interval crossing zero, indicating near-chance reliability. \textsc{Purity} falls in an intermediate range, with \texttt{Gemini Flash} yielding higher agreement (0.650) than \texttt{GPT4o-mini} (0.459).

At the rank level, Jaccard lift reveals a consistent pattern: \texttt{Gemini Flash} 
paraphrases produce more stable feature rankings than \texttt{GPT4o-mini} across all metrics, particularly at the top-10 threshold. \textsc{Detection} under \texttt{Gemini Flash} achieves a top-10 Jaccard of $43.76\times$, compared to $34.9\times$ for \texttt{GPT4o-mini}. \textsc{Fuzzing} again shows the weakest rank stability across both models. Taken together, these results suggest confirms that reliability varying substantially across metrics and paraphrase models.

\begin{table}[ht]
\centering
\caption{An overview of the results of changing the paraphrase model. }
\label{tab:paraphrase_exp}
\begin{tabular}{llccc}
\toprule
& \textbf{Metric }& \textbf{ICC(2,1)} & \textbf{Jaccard Top-10} & \textbf{Jaccard Top-20} \\
\midrule
\multirow{4}{*}{\texttt{GPT4o-mini}}
  & Simulation & 0.797 {[}0.685, 0.884{]} & 17.5$\times$ {[}2.8, 22.5{]}   & 9.7$\times$ {[}3.7, 12.4{]}  \\
  & Detection  & 0.868 {[}0.784, 0.916{]} & 34.9$\times$ {[}9.2, 42.9{]}   & 19.0$\times$ {[}9.3, 23.3{]} \\
  & Fuzzing    & 0.328 {[}$-$0.164, 0.679{]} & 17.5$\times$ {[}5.8, 34.9{]}  & 7.5$\times$ {[}4.5, 13.8{]}  \\
  & Purity     & 0.459 {[}0.333, 0.601{]} & 28.2$\times$ {[}9.2, 34.9{]}   & 13.8$\times$ {[}6.4, 23.3{]} \\
\midrule
\multirow{4}{*}{\texttt{Gemini Flash}}
  & Simulation & 0.860 {[}0.783, 0.930{]} & 43.76$\times$ {[}33.39, 49.80{]} & 15.47$\times$ {[}11.16, 19.73{]} \\
  & Detection  & 0.824 {[}0.699, 0.895{]} & 43.76$\times$ {[}35.29, 49.80{]} & 21.72$\times$ {[}19.67, 24.40{]} \\
  & Fuzzing    & 0.240 {[}0.034, 0.498{]} & 12.07$\times$ {[}7.59, 22.78{]}  & 7.17$\times$ {[}5.61, 10.50{]}  \\
  & Purity     & 0.650 {[}0.523, 0.753{]} & 30.20$\times$ {[}15.15, 38.23{]} & 19.03$\times$ {[}15.34, 22.46{]} \\
\bottomrule
\end{tabular}
\end{table}

\section{Qualitative Examples}
\label{app:qual_examples}

\subsection{Exp.1 examples - Corpus}

\paragraph{Feature 3070 on Pythia160M}
Simulation scores drop on the Wikipedia corpus  (range $= 0.369$), accompanied by a qualitative shift in the generated explanation from product and service descriptions on the Pile, C4, and Code corpora to disclosures and directives on Wikipedia. Despite moderate overall explanation similarity (cosine $= 0.831$, BERTScore 
$= 0.833$), the Wikipedia corpus retrieves a sufficiently different set 
of activating examples to produce a divergent explanation and substantially 
lower scores across simulation and purity.

\begin{tcolorbox}[
  breakable,
  colback=red!5,
  colframe=red!30,
  title={\small \textbf{Example: Corpus instability (Feature 3070, 
  BatchTopK, Pythia-160M)}},
  fonttitle=\small,
  left=8pt, right=8pt, top=6pt, bottom=6pt
]
\small

\noindent\textbf{Pile} \quad ``Descriptions of products, services, or 
technical processes'' \\
\hfill Sim: $0.813$ \quad Det: $0.650$ \quad 
Fuzz: $0.550$ \quad Pur: $0.639$

\smallskip
\noindent\textbf{C4} \quad ``Promotional or descriptive text for 
businesses, products'' \\
\hfill Sim: $0.741$ \quad Det: $0.650$ \quad 
Fuzz: $0.500$ \quad Pur: $0.679$

\smallskip
\noindent\textbf{Wikipedia} \quad ``Disclosures, revelations, or 
directives'' \\
\hfill Sim: $0.444$ \quad Det: $0.550$ \quad 
Fuzz: $0.550$ \quad Pur: $0.417$

\smallskip
\noindent\textbf{Code} \quad ``Assembly language comments, often related 
to low-level'' \\
\hfill Sim: $0.743$ \quad Det: $0.650$ \quad 
Fuzz: $0.550$ \quad Pur: $0.639$

\end{tcolorbox}

\paragraph{Feature 7223 on Pythia160M}

Simulation scores remain stable across all four corpora (range $= 0.187$) despite a notable semantic shift on the Code corpus, where the explainer describes computer viruses rather than biological pathogens (cosine $= 0.864$, BERTScore $= 0.840$). That the scorer assigns similar simulation scores to both interpretations suggests the feature captures 
a broader concept of ``virus'' or ``infection'' that spans both domains and that this conceptual stability is recoverable across corpora even when the explanation surface form diverges.

\begin{tcolorbox}[
  breakable,
  colback=red!5,
  colframe=red!30,
  title={\small \textbf{Example: Corpus stability (Feature 7223, 
  BatchTopK, Pythia-160M)}},
  fonttitle=\small,
  left=8pt, right=8pt, top=6pt, bottom=6pt
]
\small

\noindent\textbf{Pile} \quad ``Specific bacterial species, often in the context of infection'' \\
\hfill Sim: $0.874$ \quad Det: $0.600$ \quad 
Fuzz: $0.550$ \quad Pur: $0.639$

\smallskip
\noindent\textbf{C4} \quad ``Mentions of specific diseases, viruses, bacteria, or\ldots'' \\
\hfill Sim: $0.881$ \quad Det: $0.650$ \quad 
Fuzz: $0.550$ \quad Pur: $0.679$

\smallskip
\noindent\textbf{Wikipedia} \quad ``Diseases, infections, and medical conditions'' \\
\hfill Sim: $0.694$ \quad Det: $0.550$ \quad 
Fuzz: $0.550$ \quad Pur: $0.583$

\smallskip
\noindent\textbf{Code} \quad ``Descriptions or source code comments pertaining to computer viruses, often including their names, authors, or technical details'' \\
\hfill Sim: $0.793$ \quad Det: $0.650$ 
\quad Fuzz: $0.600$ \quad Pur: $0.679$

\end{tcolorbox}

\paragraph{Feature 63699 on Apertus-8B}

Despite divergent explanations across corpora, ranging 
from headings and titles on the Pile to assembly language on Code, 
simulation scores remain stable (range $= 0.146$, cosine $= 0.836$, 
BERTScore $= 0.828$). Detection and fuzzing are similarly consistent 
throughout. This feature illustrates that corpus explanation 
divergence does not necessarily produce score instability: the scorer 
assigns comparable scores to qualitatively different descriptions of 
the same underlying feature.

\begin{tcolorbox}[
  breakable,
  colback=red!5,
  colframe=red!30,
  title={\small \textbf{Example: Corpus stability (Feature 63699, 
  BatchTopK, Apertus-8B)}},
  fonttitle=\small,
  left=8pt, right=8pt, top=6pt, bottom=6pt
]
\small

\noindent\textbf{Pile} \quad ``Headings, titles, or introductory 
sections'' \\
\hfill Sim: $0.677$ \quad Det: $0.600$ \quad 
Fuzz: $0.550$ \quad Pur: $0.533$

\smallskip
\noindent\textbf{C4} \quad ``Descriptions of physical properties, 
processes, or characteristics of objects'' \\
\hfill Sim: $0.773$ 
\quad Det: $0.600$ \quad Fuzz: $0.550$ \quad Pur: $0.533$

\smallskip
\noindent\textbf{Wikipedia} \quad ``Attribution of statements, 
observations, or beliefs '' \\
\hfill Sim: $0.823$ \quad 
Det: $0.600$ \quad Fuzz: $0.500$ \quad Pur: $0.639$

\smallskip
\noindent\textbf{Code} \quad ``Assembly language code, including 
comments (\texttt{;} or\ldots)'' \\
\hfill Sim: $0.822$ \quad 
Det: $0.650$ \quad Fuzz: $0.550$ \quad Pur: $0.679$

\end{tcolorbox}

\paragraph{Feature 4081 on Apertus-8B}

Simulation scores vary substantially across corpora (range 
$= 0.327$), accompanied by qualitatively divergent explanations that share 
no obvious common concept, ranging from titles and headings on the Pile 
to passive voice constructions on Wikipedia and assembly language on Code 
(cosine $= 0.814$, BERTScore $= 0.836$). Purity varies most dramatically, dropping to $0.250$ on C4.

\begin{tcolorbox}[
  breakable,
  colback=red!5,
  colframe=red!30,
  title={\small \textbf{Example: Corpus instability (Feature 4081, 
  BatchTopK, Apertus-8B)}},
  fonttitle=\small,
  left=8pt, right=8pt, top=6pt, bottom=6pt
]
\small

\noindent\textbf{Pile} \quad ``Titles, headings, or introductory 
sentences that summarize'' \\
\hfill Sim: $0.675$ \quad Det: $0.600$ 
\quad Fuzz: $0.500$ \quad Pur: $0.533$

\smallskip
\noindent\textbf{C4} \quad ``This feature activates at the end of a 
phrase''  \\
\hfill Sim: $0.811$ \quad Det: $0.650$ \quad Fuzz: $0.500$ 
\quad Pur: $0.250$

\smallskip
\noindent\textbf{Wikipedia} \quad ``Passive voice constructions where 
an action is performed `by'''  \\
\hfill Sim: $0.636$ \quad Det: $0.600$ 
\quad Fuzz: $0.500$ \quad Pur: $0.533$

\smallskip
\noindent\textbf{Code} \quad ``Assembly language source code, 
particularly comments'' \\
\hfill Sim: $0.484$ \quad Det: $0.600$ 
\quad Fuzz: $0.500$ \quad Pur: $0.325$

\end{tcolorbox}

\subsection{Exp. 2a examples - Draws}

\paragraph{Feature 13746 on Pythia160M}

Across five independent draws from the same corpus, the explainer 
consistently identifies this feature as related to formal or technical writing (cosine similarity $= 0.883$). Yet simulation scores range from $-0.316$ to $0.707$, which is a large spread of over one unit. Detection and fuzzing remain stable throughout (variance $< 0.003$), while purity varies substantially (variance $= 0.038$).

\begin{tcolorbox}[
  breakable,
  colback=orange!5,
  colframe=orange!30,
  title={\small \textbf{Exp. 2a, Feature 13746,
  BatchTopK, Pythia-160M)}},
  fonttitle=\small,
  left=8pt, right=8pt, top=6pt, bottom=6pt
]
\small

\noindent\textbf{Draw 0} \quad ``Academic paper sections like \textit{Abstract} 
and \textit{Introduction}'' \\
\hfill Sim: $-0.316$ \quad Det: $0.600$ \quad 
Fuzz: $0.500$ \quad Pur: $0.778$

\smallskip
\noindent\textbf{Draw 1} \quad ``Formal or technical documents, including 
scientific paper abstracts, legal'' \\
\hfill Sim: $0.401$ \quad 
Det: $0.650$ \quad Fuzz: $0.500$ \quad Pur: $0.250$

\smallskip
\noindent\textbf{Draw 2} \quad ``Technical questions or discussions, often 
from forums'' \\
\hfill Sim: $0.707$ \quad Det: $0.600$ \quad Fuzz: $0.550$ 
\quad Pur: $0.556$

\smallskip
\noindent\textbf{Draw 3} \quad ``Formal or technical documents, such as 
academic papers'' \\
\hfill Sim: $0.451$ \quad Det: $0.500$ \quad Fuzz: $0.550$ 
\quad Pur: $0.476$

\smallskip
\noindent\textbf{Draw 4} \quad ``Technical discussions, problem-solving, or 
academic content'' \\
\hfill Sim: $-0.155$ \quad Det: $0.650$ \quad Fuzz: $0.550$ 
\quad Pur: $0.276$

\end{tcolorbox}

\paragraph{Feature 2208 on Pythia160M}

This feature produces the highest explanation similarity of any 
example in this section (cosine $= 0.964$, BERTScore $= 0.944$): all five 
draws generate near-identical descriptions of algebraic equations and 
variable assignments. Yet simulation scores span a range of $0.906$ --- 
from $-0.177$ to $0.729$, while detection remains fixed at $0.650$ 
throughout. This is an illustration in our data of the 
dissociation between explanation similarity and score stability: 
explanations that are essentially paraphrases of each other produce 
simulation scores that vary by nearly the full metric range.

\begin{tcolorbox}[
  breakable,
  colback=orange!5,
  colframe=orange!30,
  title={\small \textbf{Example: Sampling instability (Feature 2208, 
  BatchTopK, Pythia160M)}},
  fonttitle=\small,
  left=8pt, right=8pt, top=6pt, bottom=6pt
]
\small

\noindent\textbf{Draw 0} \quad ``Algebraic equations, variable 
assignments, and mathematical expressions'' \\
\hfill Sim: $+0.251$ 
\quad Det: $0.650$ \quad Fuzz: $0.550$ \quad Pur: $0.395$

\smallskip
\noindent\textbf{Draw 1} \quad ``Algebraic equations, variable 
assignments, and mathematical expressions'' \\
\hfill Sim: $+0.729$ \quad 
Det: $0.650$ \quad Fuzz: $0.600$ \quad Pur: $0.793$

\smallskip
\noindent\textbf{Draw 2} \quad ``Algebraic equations, variable 
assignments, and mathematical problem'' \\
\hfill Sim: $+0.498$ 
\quad Det: $0.650$ \quad Fuzz: $0.550$ \quad Pur: $0.443$

\smallskip
\noindent\textbf{Draw 3} \quad ``Algebraic equations, variable 
assignments, and mathematical expressions'' \\
\hfill Sim: $-0.008$ 
\quad Det: $0.650$ \quad Fuzz: $0.550$ \quad Pur: $0.415$

\smallskip
\noindent\textbf{Draw 4} \quad ``Mathematical equations and algebraic 
expressions, often involving multiple variables'' \\
\hfill 
Sim: $-0.177$ \quad Det: $0.650$ \quad Fuzz: $0.550$ \quad Pur: $0.303$

\end{tcolorbox}

\paragraph{Feature 45056 on Apertus-8B }

Simulation scores span the full metric range across five draws 
(range $= 1.001$, from $-0.472$ to $+0.529$), with explanations that 
shift between related but distinct concepts, such as proper nouns, biographical information, names and titles, probably depending on which examples are sampled (cosine $= 0.869$, BERTScore $= 0.846$). Detection and fuzzing are relatively stable, though detection drops to $0.500$ on Draw 2, indicating chance-level sequence classification. Purity varies substantially (range $= 0.389$), suggesting that the semantic coherence of activating examples itself depends on which examples are drawn. 

\begin{tcolorbox}[
  breakable,
  colback=orange!5,
  colframe=orange!30,
  title={\small \textbf{Example: Sampling instability (Feature 45056, 
  BatchTopK, Apertus-8B)}},
  fonttitle=\small,
  left=8pt, right=8pt, top=6pt, bottom=6pt
]
\small

\noindent\textbf{Draw 0} \quad ``Proper nouns, often capitalized, 
appearing at the beginning'' \\
\hfill Sim: $-0.125$ \quad 
Det: $0.600$ \quad Fuzz: $0.550$ \quad Pur: $0.411$

\smallskip
\noindent\textbf{Draw 1} \quad ``Biographical information, often 
including names'' \\
\hfill Sim: $+0.166$ \quad Det: $0.600$ \quad 
Fuzz: $0.550$ \quad Pur: $0.700$

\smallskip
\noindent\textbf{Draw 2} \quad ``Specific names or titles followed by 
a brief description '' \\
\hfill Sim: $+0.529$ \quad Det: $0.500$ 
\quad Fuzz: $0.500$ \quad Pur: $0.333$

\smallskip
\noindent\textbf{Draw 3} \quad ``Proper nouns, often names of people, 
places'' \\
\hfill Sim: $+0.148$ \quad Det: $0.650$ \quad 
Fuzz: $0.500$ \quad Pur: $0.525$

\smallskip
\noindent\textbf{Draw 4} \quad ``Specific names, places, or technical 
terms'' \\
\hfill Sim: $-0.472$ \quad Det: $0.650$ \quad 
Fuzz: $0.550$ \quad Pur: $0.318$
\end{tcolorbox}

\newpage
\section*{NeurIPS Paper Checklist}

The checklist is designed to encourage best practices for responsible machine learning research, addressing issues of reproducibility, transparency, research ethics, and societal impact. Do not remove the checklist: {\bf The papers not including the checklist will be desk rejected.} The checklist should follow the references and follow the (optional) supplemental material.  The checklist does NOT count towards the page
limit. 

Please read the checklist guidelines carefully for information on how to answer these questions. For each question in the checklist:
\begin{itemize}
    \item You should answer \answerYes{}, \answerNo{}, or \answerNA{}.
    \item \answerNA{} means either that the question is Not Applicable for that particular paper or the relevant information is Not Available.
    \item Please provide a short (1--2 sentence) justification right after your answer (even for \answerNA). 
\end{itemize}

{\bf The checklist answers are an integral part of your paper submission.} They are visible to the reviewers, area chairs, senior area chairs, and ethics reviewers. You will also be asked to include it (after eventual revisions) with the final version of your paper, and its final version will be published with the paper.

The reviewers of your paper will be asked to use the checklist as one of the factors in their evaluation. While \answerYes{} is generally preferable to \answerNo{}, it is perfectly acceptable to answer \answerNo{} provided a proper justification is given (e.g., error bars are not reported because it would be too computationally expensive'' or ``we were unable to find the license for the dataset we used''). In general, answering \answerNo{} or \answerNA{} is not grounds for rejection. While the questions are phrased in a binary way, we acknowledge that the true answer is often more nuanced, so please just use your best judgment and write a justification to elaborate. All supporting evidence can appear either in the main paper or the supplemental material, provided in appendix. If you answer \answerYes{} to a question, in the justification please point to the section(s) where related material for the question can be found.

IMPORTANT, please:
\begin{itemize}
    \item {\bf Delete this instruction block, but keep the section heading ``NeurIPS Paper Checklist"},
    \item  {\bf Keep the checklist subsection headings, questions/answers and guidelines below.}
    \item {\bf Do not modify the questions and only use the provided macros for your answers}.
\end{itemize}


\begin{enumerate}

\item {\bf Claims}
    \item[] Question: Do the main claims made in the abstract and introduction accurately reflect the paper's contributions and scope?
    \item[] Answer: \answerYes{}
    \item[] Justification: We put the claims made by this paper in the abstract and introduction

\item {\bf Limitations}
    \item[] Question: Does the paper discuss the limitations of the work performed by the authors?
    \item[] Answer: \answerYes{}
    \item[] Justification: We discuss the limitations in \autoref{sec:conclusion}.

\item {\bf Theory assumptions and proofs}
    \item[] Question: For each theoretical result, does the paper provide the full set of assumptions and a complete (and correct) proof?
    \item[] Answer: \answerNA{}
    \item[] Justification: This paper does not include new theoretical results

    \item {\bf Experimental result reproducibility}
    \item[] Question: Does the paper fully disclose all the information needed to reproduce the main experimental results of the paper to the extent that it affects the main claims and/or conclusions of the paper (regardless of whether the code and data are provided or not)?
    \item[] Answer: \answerYes{}
    \item[] Justification: We provide implementation details in \autoref{sec:preliminaries}, with more details in \autoref{app:experiments}, \autoref{app:instability_quant} and \autoref{app:sae_training_details}.

\item {\bf Open access to data and code}
    \item[] Question: Does the paper provide open access to the data and code, with sufficient instructions to faithfully reproduce the main experimental results, as described in supplemental material?
    \item[] Answer: \answerYes{}
    \item[] Justification: We have provided an anonymous repository link to reproduce the code and the results with Pythia-160M. We cannot share the link of the pre-trained SAEs on Apertus-8B, due to
    conflict with anonymity. 

\item {\bf Experimental setting/details}
    \item[] Question: Does the paper specify all the training and test details (e.g., data splits, hyperparameters, how they were chosen, type of optimizer) necessary to understand the results?
    \item[] Answer: \answerYes{}
    \item[] Justification: Information can be found in \autoref{sec:preliminaries} and \autoref{sec:insights}, with more details on the models in \autoref{app:sae_training_details}.

\item {\bf Experiment statistical significance}
    \item[] Question: Does the paper report error bars suitably and correctly defined or other appropriate information about the statistical significance of the experiments?
    \item[] Answer: \answerYes{}
    \item[] Justification: Confidence intervals are included for ICC results, see \autoref{tab:icc}. 

\item {\bf Experiments compute resources}
    \item[] Question: For each experiment, does the paper provide sufficient information on the computer resources (type of compute workers, memory, time of execution) needed to reproduce the experiments?
    \item[] Answer: \answerYes{}
    \item[] Justification: We included an estimation of LLM calls and time in \autoref{app:compute}.
    
\item {\bf Code of ethics}
    \item[] Question: Does the research conducted in the paper conform, in every respect, with the NeurIPS Code of Ethics \url{https://neurips.cc/public/EthicsGuidelines}?
    \item[] Answer: \answerYes{}
    \item[] Justification: The research in the paper conforms with the NeurIPS code of Ethics. 

\item {\bf Broader impacts}
    \item[] Question: Does the paper discuss both potential positive societal impacts and negative societal impacts of the work performed?
    \item[] Answer: \answerNA{}.
    \item[] Justification: This work is a methodological study of evaluation reliability and introduces no new models or deployment systems. It has no direct path to harmful applications, with the note that any potential societal impact is indirect, through its contribution to more reliable interpretability evaluation practices.
    
\item {\bf Safeguards}
    \item[] Question: Does the paper describe safeguards that have been put in place for responsible release of data or models that have a high risk for misuse (e.g., pre-trained language models, image generators, or scraped datasets)?
    \item[] Answer: \answerNA{}.
    \item[] Justification: This paper evaluates reliability of an existing method, and does not pose such risk. 

\item {\bf Licenses for existing assets}
    \item[] Question: Are the creators or original owners of assets (e.g., code, data, models), used in the paper, properly credited and are the license and terms of use explicitly mentioned and properly respected?
    \item[] Answer: \answerYes{}
    \item[] Justification: See \autoref{tab:corpora} and \autoref{app:pythiadetails}. All owners of models and data are properly cited. 

\item {\bf New assets}
    \item[] Question: Are new assets introduced in the paper well documented and is the documentation provided alongside the assets?
    \item[] Answer: \answerYes{}
    \item[] Justification: We document all assets. 

\item {\bf Crowdsourcing and research with human subjects}
    \item[] Question: For crowdsourcing experiments and research with human subjects, does the paper include the full text of instructions given to participants and screenshots, if applicable, as well as details about compensation (if any)? 
    \item[] Answer: \answerNA{}.
    \item[] Justification: There is no crowdsourcing nor research with human subjects. 

\item {\bf Institutional review board (IRB) approvals or equivalent for research with human subjects}
    \item[] Question: Does the paper describe potential risks incurred by study participants, whether such risks were disclosed to the subjects, and whether Institutional Review Board (IRB) approvals (or an equivalent approval/review based on the requirements of your country or institution) were obtained?
    \item[] Answer: \answerNA{}.
    \item[] Justification: There is no crowdsourcing nor research with human subjects. 

\item {\bf Declaration of LLM usage}
    \item[] Question: Does the paper describe the usage of LLMs if it is an important, original, or non-standard component of the core methods in this research? Note that if the LLM is used only for writing, editing, or formatting purposes and does \emph{not} impact the core methodology, scientific rigor, or originality of the research, declaration is not required.
    \item[] Answer:\answerYes{}
    \item[] Justification: LLMs are part of the pipeline we evaluate. The pipeline with the LLMs is shown in \autoref{fig:overview_auto}. Usage of the LLMs is described in \autoref{sec:preliminaries}.

\end{enumerate}

\end{document}